\documentclass[final,5p,times,twocolumn,authoryear,nopreprintline]{elsarticle}

\usepackage{amssymb}

\usepackage[table, dvipsnames]{xcolor}
\usepackage{soul}
\usepackage{amsmath}
\usepackage{algorithm}
\usepackage[noend]{algpseudocode}
\usepackage{tabularx}
\usepackage{caption}
\usepackage{subcaption}
\usepackage{multirow}
\usepackage{dblfloatfix}
\usepackage{hyperref}
\usepackage{pgfplots}
\pgfplotsset{compat=1.17}

\journal{Elsevier}

\begin{document}

\begin{frontmatter}



\title{One Class One Click: Quasi Scene-level Weakly Supervised Point Cloud Semantic Segmentation with Active Learning}


\author[polyu]{Puzuo Wang}
\ead{puzuo.wang@connect.polyu.hk}

\author[polyu,polyusri,polyuscri]{Wei Yao\corref{cor1}}
\ead{wei.hn.yao@polyu.edu.hk}

\author[polyu]{Jie Shao}

\address[polyu]{Dept. of Land Surveying and Geo-Informatics, The Hong Kong Polytechnic University, Hung Hom, Kowloon, Hong Kong}
\address[polyusri]{The Hong Kong Polytechnic University Shenzhen Research Institute, Shenzhen, China}
\address[polyuscri]{Otto Poon Charitable Foundation Smart Cities Research Institute, The Hong Kong Polytechnic University, Hung Hom, Hong Kong}

\cortext[cor1]{Corresponding author.}


\begin{abstract}
Reliance on vast annotations to achieve leading performance severely restricts the practicality of large-scale point cloud semantic segmentation. For the purpose of reducing data annotation costs, effective labeling schemes are developed and contribute to attaining competitive results under weak supervision strategy. Revisiting current weak label forms, we introduce One Class One Click (OCOC), a low cost yet informative quasi scene-level label, which encapsulates both point-level and scene-level annotations. An active weakly supervised framework is proposed to leverage scarce labels by involving weak supervision from both global and local perspectives. Contextual constraints are imposed by an auxiliary scene classification task, respectively based on global feature embedding and point-wise prediction aggregation, which restricts the model prediction merely to OCOC labels within a sub-cloud. Furthermore, we design a context-aware pseudo labeling strategy, which effectively supplement point-level supervisory signals subject to OCOC labels. Finally, an active learning scheme with a uncertainty measure - temporal output discrepancy is integrated to examine informative samples and provides guidance on sub-clouds query, which is conducive to quickly attaining desirable OCOC annotations and reduces the labeling cost to an extremely low extent. Extensive experimental analysis using three LiDAR benchmarks respectively collected from airborne, mobile and ground platforms demonstrates that our proposed method achieves very promising results though subject to scarce labels. It considerably outperforms genuine scene-level weakly supervised methods by up to 25\% in terms of average F1 score and achieves competitive results against full supervision schemes. On terrestrial LiDAR dataset - Semantics3D, using approximately 2\textpertenthousand{} of labels, our method achieves an average F1 score of 85.2\%, which increases by 11.58\% compared to the baseline model.

\end{abstract}




\end{frontmatter}


\section{Introduction}
\label{sec:intro}

LiDAR point clouds depict precise three-dimensional (3D) representation of real-world scenes, providing valuable geospatial and geometric-structural clues for various remote sensing tasks \citep{YAO2011260,POLEWSKI2015252}. Among these applications, a fundamental processing is to acquire point-wise semantics, which is regarded as point cloud semantic segmentation or classification.

In recent years, we have seen incredible advancements in point cloud semantic segmentation along with the success of deep learning. Benefiting from advanced network architecture design, emerging approaches have continuously achieved and surpassed state-of-the-art results \citep{pointnet,kpconv,HUANG202062,pointtransformer}. However, most of them rely on large amounts of well labeled training samples, which is referred to as data hungry issue. Typically, labeling work is associated with heavy workloads, even for experienced operators. Besides, the irregular discrete distribution and 3D structure of point clouds dramatically increases the difficulty level of interpretation.

While data labeling is a difficult and time-consuming job, the generation and collection of raw point clouds has become simple and convenient owing to the advances in LiDAR technology and diversified data acquisition platforms. Currently, millions of ultra dense points could be captured within a short time period. Given massive point clouds to be classified, an intuitive idea is whether promising results can be attained without the necessity to label the entire scene as training samples. The workload of data annotation will be significantly reduced if a comparable performance is achieved using only scarce labels, which significantly contributes to the efficacy and practicality of real-life applications. In fact, modern deep models tend to maintain satisfactory performance when largely decreasing annotation abundance. Experimental results from \citet{sqn} indicate that there is merely a slight accuracy degradation even if only using 1\% of sparsely distributed labels. In this study, we further explore novel solutions under the label scarcity issue.

Weakly supervised learning are proposed to address situations of incomplete annotations. We tackle the problem only with inductive scheme in this study. Among most of weakly supervised methods, extra information  were created or predicted for performance boost in addition to original weak labels. Many strategies are proposed to involve contrastive and consistency constraints for loss calculation, such as data augmentation \citep{WS-DAN}, temporal consistency \citep{temporalensenbling}, and model parameter consistency \citep{meanteachers}. On the other hand, a branch of methods aimed to creating extra supervisory signals, such as virtual sample \citep{zhang2018mixup}, pseudo label \citep{lee2013pseudo}, and label propagation \citep{Iscen_2019_CVPR}. For weakly supervised point cloud semantic segmentation, methods often leveraged point cloud characteristics to address the problem of scarce annotations. For instance, \citet{xu2020weakly} proposed a Siamese self-supervision by rotation and mirror flipping  and preserved local semantic smoothness based on spatial and color manifold. In \citet{SSPC-Net}, a dynamic superpoint-level label propagation was conducted progressively to generate pseudo labels. \citet{Hou_2021_CVPR} considered contrastive scene context in a 3D pre-training framework that utilized both point-level correspondences and spatial contexts.

\begin{figure}[t]
\centering
\includegraphics[width=1\linewidth]{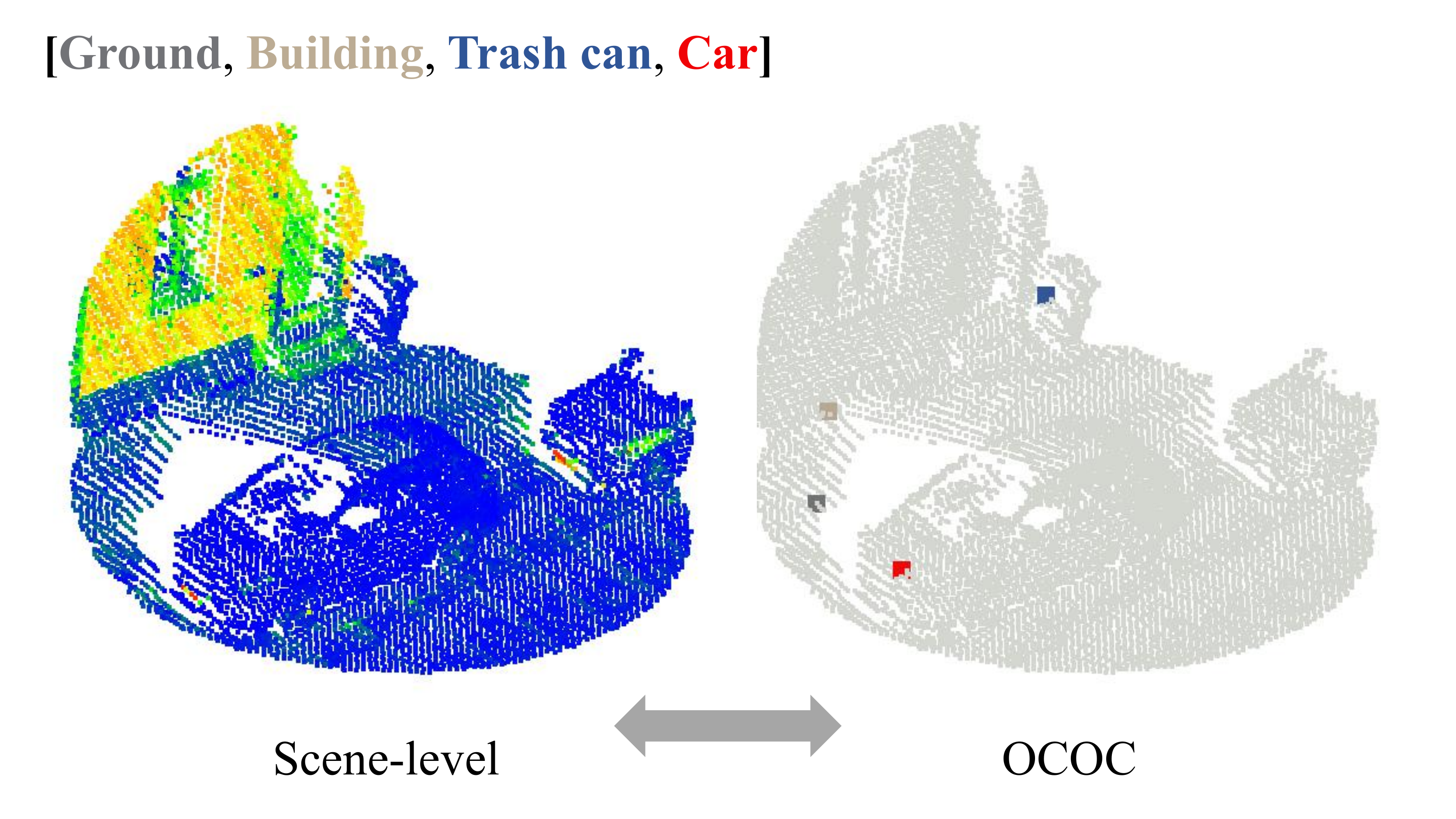}
\caption{Comparison between scene-level and point-level (weak) labels.}
\label{fig:wl}
\end{figure}

How to organize the weak label in an efficient format is essential issue in a effort to reduce labeling costs and secure satisfactory results simultaneously, definitely affecting the design of the weakly supervised methods. In point cloud processing, sparsely distributed point-level labels were commonly used \citep{WANG2022237,Zhang_Li_2021,9454396}, which can be directly integrated with segmentation networks without modification. Scene-level weak labels were considered in some works \citep{wei2020multi,LIN202279}, in which existing categories in a sub-cloud are predefined. Moreover, 3D bounding box is utilized to build the bridge between semantic segmentation and object detection \citep{Box2Seg,Box2Mask}. However, studies based on box-level labels were currently limited to in-door environments. Scene-level labels seem to be more accessible than point-level ones, but also more challenging to work with, which often results in poor performance. By revisiting the labeling process of scene-level annotation, we argue that the scene-level label can be transformed to point-level labels while maintaining similar labeling costs. As shown in Fig.~\ref{fig:wl}, to acquire the scene-level label, it is necessary to interpret existent classes in the scene, which basically equals to several clicks on points of each category. To this end, we propose a quasi scene-level weakly supervised framework, named as One Class One Click (OCOC). Each existent category in a sub-cloud is assigned with only one point-level annotation. It enables to train a conventional semantic segmentation network while scene-level contextual information is obtained.

Identifying most informative weak labels is also important for weak supervision, which could maximize the model performance with high training efficiency under a fixed labeling budget. Currently, most of studies adopted randomly initialization, without considering label information correlation. \citet{annals_wang2022} explored weak label distribution issue and found that using class-balanced weak labels achieved better performance using deep learning based methods. However, how to specifically locate the labeled points was not further investigated. Active learning enables human operators interactively to query and annotate desirable samples, which could lead to a highly efficient training process. Typical active learning methods cyclically infer informative samples prone to be misclassified and assign corresponding labels, boosting model performance in an iterative training manner \citep{al_survey}. Moreover, active labeling shows significant potential in transferring current model knowledge to unseen scenes. In light of training efficiency, batch-based query strategy is often applied when combining active learning with deep neural networks. Thus, it is necessary to take into account data information and diversity simultaneously. Considering  adjacent scenes with high spatial similarity, we construct spatially diverse samples based on inference from an information measure, Temporal Output Discrepancy (TOD) \citep{Huang_2021_ICCV}, which are theoretically associated with loss function of deep models.

In this work, we combine weak supervision and active learning for point cloud semantic segmentation. We first introduce OCOC, a cost-effective quasi scene-level weak label. Then, a weakly supervised method is developed to leverage obtained labels. Apart from point-wise supervision, contextual constraints are applied to global features and point-wise predictions, respectively. Moreover, we generate context-aware pseudo labels to provide extra supervisory signals, such that which are limited to existing OCOC categories within a sub-cloud. In order to further explore desirable weak labels, we adopt active learning strategy to query and generate most informative training samples. The measure - TOD, which estimates the data loss by evaluating the discrepancy of model outputs at different training steps, is incorporated to quantify a uncertainty measure. Then, a batch of sub-clouds are collected to expand labeled pool with respect to the uncertainty and spatial diversity, and a simulation strategy is proposed to acquire corresponding OCOC by imitating manual annotation. We apply KPConv \citep{kpconv} as the backbone network. Experiments on three LiDAR benchmarks acquired from different platforms indicate that only approximately 2\textpertenthousand{} of labels are required to achieve results close to full supervision schemes. Our main contributions are summarized as follows.

\begin{itemize}[leftmargin=*]

\item[$\bullet$] We propose an active weakly supervised point cloud semantic segmentation framework, by leveraging One Class One Click (OCOC), a cost-effective quasi scene-level label which contains both contextual and point-wise semantic information. 

\item[$\bullet$] A weakly supervised method is proposed by introducing contextual constraints from OCOC simultaneously to global scene-level and local point-wise predictions. And a context-aware pseudo labeling strategy is further developed to supply additional supervisory signals.

\item[$\bullet$] We develop an active leaning scheme based on Temporal Output Discrepancy (TOD) for effectively expanding labeled pool. Considering spatial diversity, informative sub-clouds are sequentially identified for corresponding OCOC annotation under TOD guidance.

\item[$\bullet$] Evaluation of the proposed approach using three multi-platform benchmarks demonstrates that very promising performance close to full supervision schemes is achieved with extremely low labeling costs.

\end{itemize}

The rest of the study is organized as follows. In Section~\ref{sec:re}, we systematically review weakly supervised learning for point cloud semantic segmentation and deep learning based active learning. The proposed methodology is described in detail in Section~\ref{sec:method}. Section~\ref{sec:exp} presents the datasets and implementations. Regarding Section~\ref{sec:res}, we present an extensive experimental analysis to compare and analyze the effectiveness of the proposed method. The concluding remarks are provided for future work in Section~\ref{sec:con}.

\section{Related work}
\label{sec:re}

While deep learning based methods have reached quite high accuracy in a wide range of tasks, the high reliance on precise data annotations restricts its practicality in real-world applications. To solve the issue, weakly supervised methods seek to maintain superior performance with limited labels. On the other hand, active learning approaches reduce redundant labeling costs through interactions between experts and model predictions. In this section, we conduct a comprehensive review of studies involving these two streams. 

\subsection{Weakly supervised point cloud semantic segmentation}
\label{sec:re_wsss}

Currently, weakly supervised learning draws increasing attention to the field of point cloud semantic segmentation. Given incomplete and sparse annotations, weakly supervised methods create extra constraints as auxiliary information to guide model training. Based on different workflows and strategies, we categorize these methods into two main groups, contrastive constraint and supervisory signal expansion. It should be noted that a hybrid framework that combines two mechanisms is verified for achieving better performance. Here we separately summarize two categories, though a hybrid strategy is applied in most of studies.

\subsubsection{Contrastive constraint}
\label{sec:re_wsss_cl}

The concept of contrastive learning comes from a simple idea, that visual interpretation results or calculated features should keep consistency when comparing a pair of similar data. An intuitive method is to apply perturbations such as rotation and scaling to the original point cloud and minimize prediction discrepancy between two data \citep{xu2020weakly}. In \citet{Zhang_2021_ICCV}, scene-wise transformation and point-wise displacement were applied to point clouds, whereby the Jensen-Shannon divergence was adopted as the self-distillation loss. \citet{arxiv2021_wei} designed a cross-sample feature reallocating module to enhance features and proposed a consistency loss by comparison with basic ones. Apart from comparing corresponding point pairs based on data augmentation, some studies conducted the comparison between points from different spatial locations. An early study was proposed for self-supervised point cloud representation learning, in which points with spatial differences were collected as negative pairs, forced to produce distinguished features \citep{PointContrast}. For semantic segmentation, pseudo labels were considered powerful alternatives to guide positive/negative pair construction. In \citet{Jiang_2021_ICCV}, negative feature pairs with different semantic predictions were forced to be dissimilar. Moreover, HybridCR \citep{Li_2022_CVPR} developed two contrastive consistency based regularizers from local and global perspectives, which encouraged the anchor point being similar to matched positive points while being dissimilar to negative points with respect to posterior probability. While the global guidance utilized class prototypes to construct point pairs, the local constraint was implemented through comparisons between spatial neighbors.

\subsubsection{Supervisory signal expansion}
\label{sec:re_wsss_sse}

Supervisory signal expansion is a sort of intuitive strategy, that creates targets with the same format as original annotations and treats them as ground truth for training simultaneously. Pseudo labeling, which annotates a large amount of unlabeled data based on the model predictions trained on a small set of labeled data, is commonly used among diverse weakly supervised methods. Considering the enormous variation in point cloud density in autonomous driving scenarios, a class-range-balanced pseudo labeling was developed by proportionally selecting pseudo labels from divided blocks and designing thresholds in a class-wise manner \citep{Unal_2022_CVPR}. \citet{WANG2022237} proposed an online soft pseudo labeling method, in which pseudo labels were generated and updated from ensemble predictions with different weights based on entropy, enabling an efficient and parameter-free training process. In \citet{DENG202278}, pseudo label refinement was performed through relational graph constructed from local and non-local points. Additionally, by leveraging spatial structural information, superpoint or graph structure were adopted for pseudo label propagation. \citet{9811904} generated superpoints through region growing based on geometry and color. Then, according to purity of pseudo labels in the superpoint, all contained points were assigned with either dominant label or no label. In \citep{Liu_2021_CVPR}, objects contained in supervoxels with initial weak labels were assigned with pseudo labels for self-training, and pseudo labels were propagated through a graph structure based on geometric homogeneity of supervoxels. Similarly, a dynamic label propagation strategy was proposed by \citep{SSPC-Net} to progressively propagate superpoint-level pseudo labels from initial weak labels and dynamically select pseudo labels with a superpoint dropout strategy. On the other hand, some studies leveraged non-point-level weak labels, by transforming them to point-level ones during training. Box2Seg \citep{Box2Seg} aimed to learn dense semantics of point clouds with 3D bounding box level annotations. A unsupervised 3D GrabCut algorithm was first introduced to extract foreground points for generating pseudo labels. Based on scene-level labels, \citet{LIN202279} proposed a two-step strategy, in which initial results generated by a point class activation map were regarded as pseudo labels to retrain a semantic segmentation network in the second step. 

Among weak supervised methods, effective weak label format largely determines the success of the proposed strategies for the sake of saving labeling costs. To cope with large-scale point cloud semantic segmentation, we introduce OCOC, which enables an intuitive yet effective weak supervision method with extremely low labeling costs.

\subsection{Active learning on point clouds }
\label{sec:re_al}

Alongside model training, active learning methods incrementally expand the labeled pool by interactively allowing a user to be asked about the label of certain instances that are currently unlabeled. This is the promise of active learning - if the model asks smart questions, it might be able to get examples that are very informative and reach a high level of generalization accuracy with a much smaller labeled dataset than it would have if that dataset had been created using random sampling. Uncertainty sampling is probably the simplest and most straightforward idea to make the model query the example which it is least certain about. And a designed criteria calculated from model's predictions for the data over labels could be utilized as the measure. For instance, margin sampling \citep{Scheffer2001} queried data with minimal probability difference between top and second most probably predicted classes. A more general strategy used Shannon entropy \citep{shannon} as the uncertainty measure, which may be the most popular criterion. In deep learning, loss value was regarded as a perfect guidance for sample selection in active learning \citep{Yoo_2019_CVPR}. However, it cannot be calculated from unlabeled data due to lacking annotations. To incorporate with deep models, some studies aimed to explore implicit loss information during training. In \citet{Yoo_2019_CVPR}, a loss prediction module was proposed to infer approximate value. \citet{Huang_2021_ICCV} analyzed relations between loss value and model prediction discrepancy during training and proved that the discrepancy is the lower bound of real loss value. Moreover, since a batch-mode sampling strategy was necessary for deep learning models, data redundancy within a batch inspired researchers to further consider the diversity among the samples. A core-set approach was developed to select most representative samples from remaining unlabeled pool that is as diverse as possible and represents the complete data distribution \citep{sener2018active}. Through theoretical analysis, the Wasserstein distance was adopted by \citet{pmlr-v108-shui20a} for modeling the interactions in active learning as distribution matching, divulging an explicit uncertainty-diversity trade-off.

Until now, there are still handful works available for point cloud processing in the context of active learning. \citet{polewski2015active} applied a Renyi entropy guided active semi-supervised model to detect standing dead trees from airborne laser scanning data combined with infrared images. In \citet{lin2020active}, three uncertainty measures were comparatively studied for active semantic ALS point cloud segmentation, where a tile of point clouds were incrementally added to model training. Most recently, methods also attempted to combine active learning and weak supervision. \citet{Wu_2021_ICCV} actively expanded superpoint-level labels for semantic segmentation. Color difference and surface variation were considered for stimulating diversity awareness. An analogous framework with noise-aware iterative labeling strategy was also proposed in \citet{shao2022active}. \citet{Wu_2021_ICCV} and \citet{shao2022active} regarded superpoints as labeling unit without semantic ambiguity, therefore allocating each of them with only one label. Nevertheless, the segmentation process for point clouds are not error-free, inevitably leading to noisy labels.

\begin{figure*}[ht]
\centering
\includegraphics[width=1.0\linewidth]{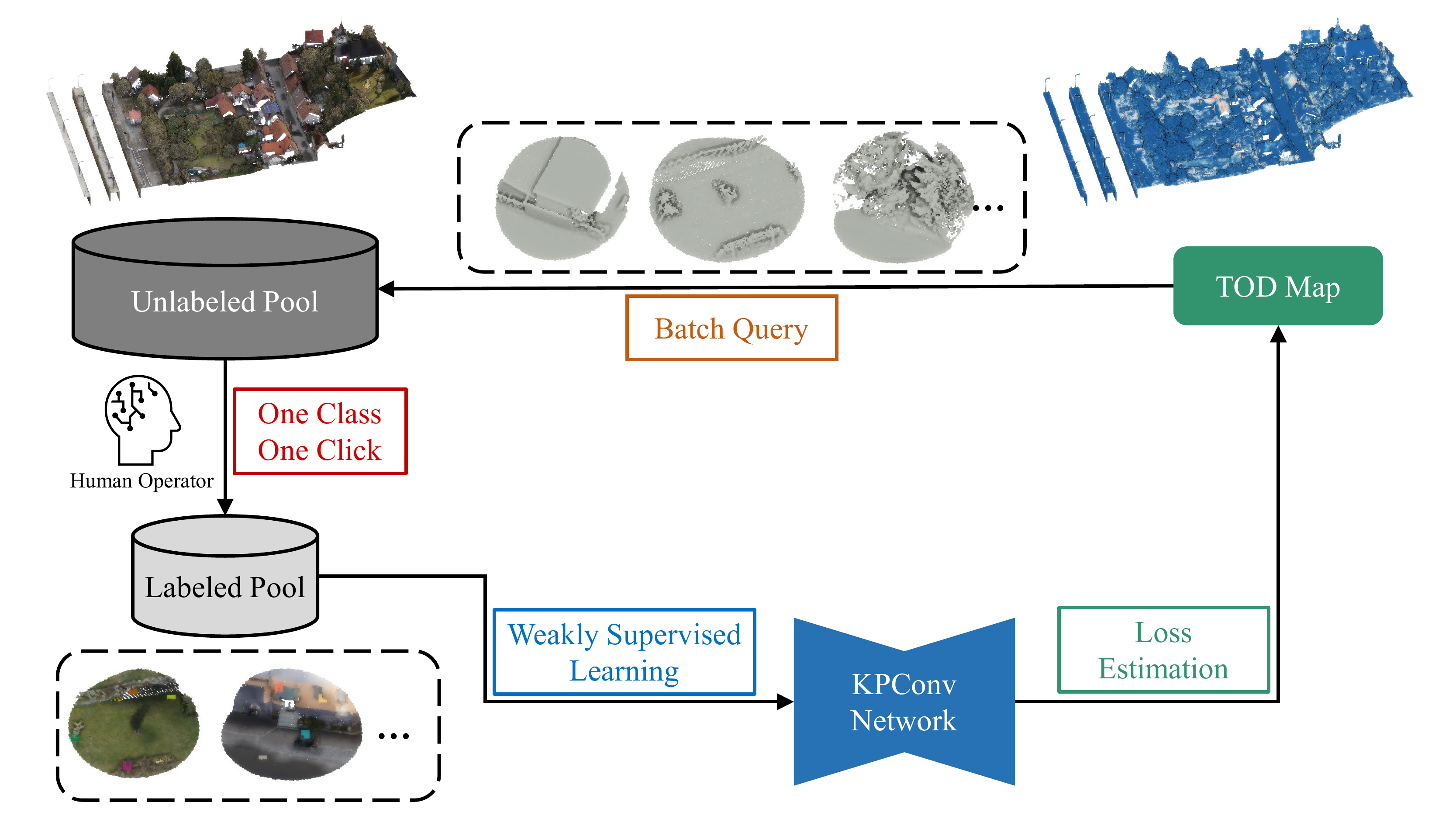}
\caption{The pipeline of proposed active weakly supervised strategy for point cloud semantic segmentation. Given raw point clouds as unlabeled pool, we first initialize labeled pool with random sub-clouds selection and OCOC annotation. Next, a weakly supervised method is developed leveraging scarce labels. Once the trained model reaches convergence, TOD map inferred from the prediction provides guidance on informative batch query for OCOC annotation. The whole framework is executed in a cyclic pattern.}
\label{fig:workflow}
\end{figure*}

\section{Methodology}
\label{sec:method}

\subsection{Overview}
\label{sec:m_oberview}

In this study, we introduce One Class One Click (OCOC) for active weakly supervised point cloud semantic segmentation. Given the point cloud $\mathbf{P}\in \mathbb{R}^{N\times D}$ consisting of $N$ points and corresponding $D$ dimensional attributes and $C$ categories, $K$ sub-clouds $\{\mathbf{s}_k\}$ $(i=1,2,...,K,\mathbf{s}_k\in \mathbb{R}^{M_k\times D})$ are sampled from $\mathbf{P}$ as training data. The corresponding weak labels $\mathbf{l}_k$ are constructed as allocating OCOC for $\mathbf{s}_k$. A weakly supervised framework is developed to leverage quasi sub-cloud level labels. 

Incomplete supervision with weak labels is used as the baseline. To exploit unlabeled data within sub-clouds, We develop semantic constraints for global feature embeddings and point-wise predictions. Since scene-level label information of a sub-cloud is contained in OCOC, we conduct multi-label scene recognition as an auxiliary task to enhance encoding layers. Max predicted probability over multiple labels that also indicates the multi-class occurrence in the sub-cloud, is utilized as the contextual constraint. Moreover, we generate context-aware pseudo labels to provide extra supervisory signals, in which pseudo labels are constrained by OCOC as well. An active learning mode is further combined, which cyclically queries most informative sub-clouds. Temporal Output Discrepancy (TOD) \citep{Huang_2021_ICCV} is adopted to estimate the point-wise uncertainty across the whole point cloud. Then, TOD is refined by spatial and semantic smoothing, and points with local maximum are selected as central point candidates for sub-clouds extraction. Finally, OCOC labeling will be conducted by the human operator under guidance of TOD values. The workflow of the proposed method is illustrated in Fig.~\ref{fig:workflow}.

\subsection{One Class One Click}
\label{sec:m_ococ}
As discussed in Sec.~\ref{sec:intro}, OCOC is defined as annotating only one point for each existing category in a sub-cloud $\mathbf{s}$. With extremely low labeling costs, OCOC acquires both point-level and scene-level label information. Here we discuss the superiority of OCOC.

\paragraph{Comparison with point-level weak label}
Point-level weak label is defined as spatially sparse annotations. While it is often collected by random, OCOC assigns point-level labels with minimum labeling costs and obtains extra scene-level information in parallel. Additionally, instead of generating training samples across the whole point cloud, only extracted sub-clouds are utilized for training under OCOC annotation, which enables a simple mechanism for sample size control and training efficiency improvement. 

\paragraph{Comparison with scene-level weak label}
Scene-level weak label denotes a textual vector comprising existent categories in a sub-cloud. On top of this, OCOC also preserves point-level labels, which enables straightly to train a standard semantic segmentation network. By contrast, a category localization strategy is usually necessary for scene-level weak label to obtain point-wise predictions, often resulting in an inaccurate semantic boundary.

\begin{figure*}[ht]
\centering
\includegraphics[width=1.0\linewidth]{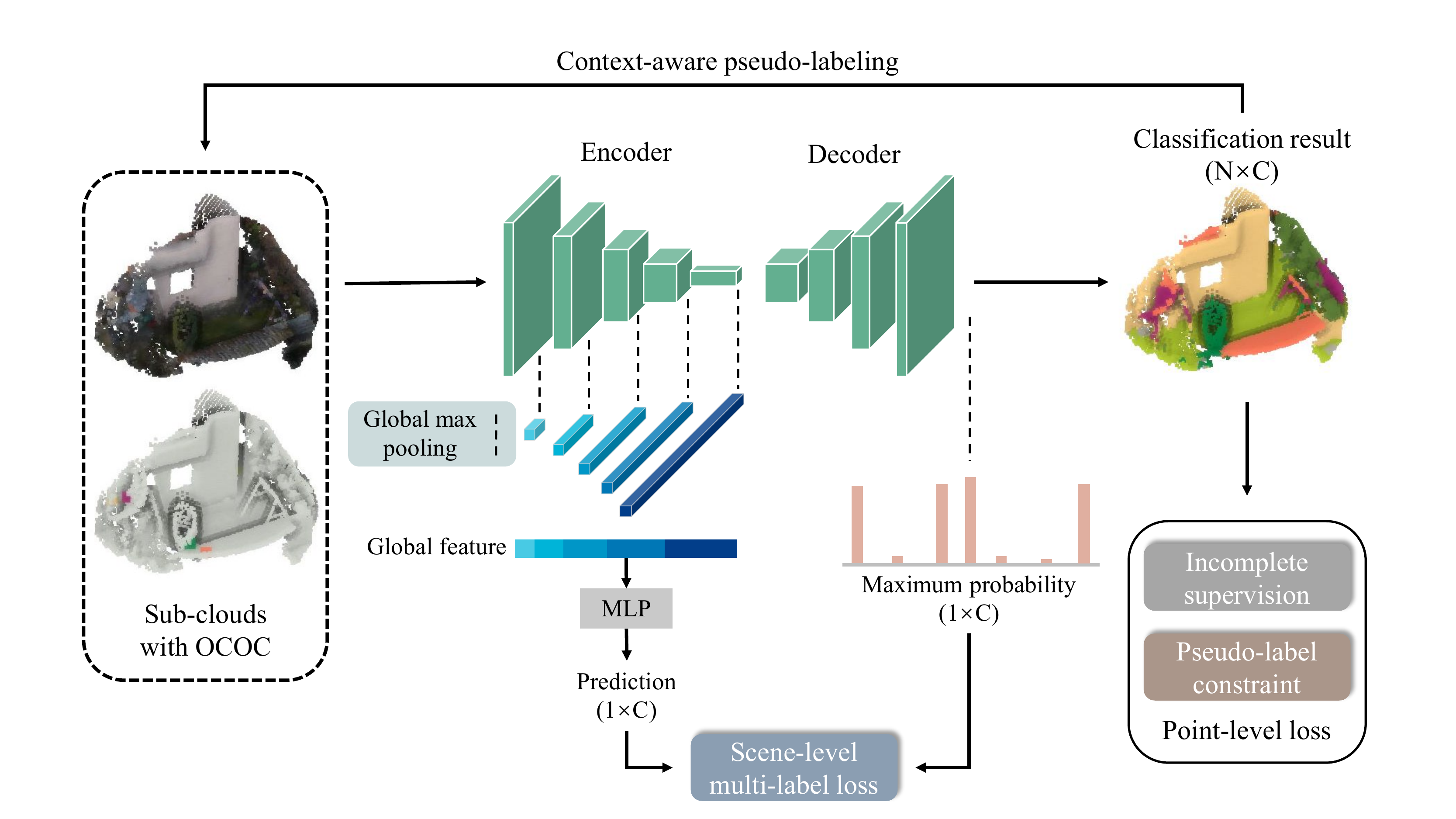}
\caption{Illustration of the proposed weakly supervised method. Global feature is generated through concatenation of global max pooled features at each encoding layer, predicting scene-level label with a MLP layer. Maximum probability of point-wise predictions with respect to semantic classes is also calculated for scene-level characterization through global max pooling, restricted by OCOC label. Along with sparse labels, context-aware pseudo labels, which are confined to existing categories in a sub-cloud, are created for point-level supervision.}
\label{fig:wsl}
\end{figure*}

\subsection{Weak supervision with contextual awareness }
\label{sec:m_ws}

We propose a weakly supervised method leveraging OCOC, contributing to training a deep model from local and global perspectives. KPConv \citep{kpconv} is chosen as the backbone network owing to its strong generalizability and superior performance, which can be also replaced by other task-oriented deep models.

\subsubsection{Point-wise incomplete learning}
\label{sec:m_ws_pts}

We first conduct incomplete learning on labeled data. Since only few points ($M_w$) in sub-clouds $\mathbf{s}$ are annotated, denoted as $\mathbf{s}_w$, we calculate the loss of these points to perform backpropagation. The softmax cross-entropy considering the class imbalance issue is applied to loss calculation:

\begin{equation}\label{equ:L_seg}
	\mathcal{L}_{\text {seg}} = -\frac{1}{\left | \mathbf{s}_w \right |} \sum_{i}^{M_w} w_{l_i}\sum_{c}^{C}y_{ic}\log p_{ic}
\end{equation}
where $p_{ic}$ stands for predicted probability of class $c$ for point $x_i$, and $y_{ic}$ = 1 if $c$ equals to label $l_i$, otherwise 0. $w$ is presented to mitigate label imbalance issue, which is calculated according to the proportion of category $c$:

\begin{equation}\label{equ:w_class}
	w_c=\frac{1}{\sqrt{M_{c}} \sum_{i=1}^{C} \frac{1}{\sqrt{M_{i}}}}
\end{equation}

\subsubsection{Scene-level constraint}
\label{sec:m_ws_scene}

Contextual information has been proved beneficial to point cloud semantic segmentation \citep{9756643}, which is utilized in this study for weak supervision. Inspired from scene recognition task, we predict scene-level labels of sub-clouds, which serves as an auxiliary task to enhance encoding layers. A multi-scale feature aggregation module is developed for capturing and learning comprehensive feature representation. Specifically, after extracting latent features $f^l$ of each encoding layer $l$, global max pooled feature across spatial dimension $\bar{f}^l$ is extracted, where $\bar{f}^l=\max_{i}f_{i}^l$. Then, $\bar{f}^l$ from different layers are aggregated by concatenation, being fed into a multilayer perceptron (MLP) layer to produce scene-level semantic feature, defined as 

\begin{equation}\label{equ:mlp}
	\bar{g}=\text{MLP}(\text{concat}[\bar{f}^1, \bar{f}^2, ..., \bar{f}^L])
\end{equation}
Sigmoid function is utilized to determine scene-level prediction, formulated as

\begin{equation}\label{equ:sigmoid}
	\bar{z}=\log \frac{1}{1+\exp(-\bar{g})}
\end{equation}
We adopt Binary Cross Entropy (BCE) to calculate scene-level multi-label loss, written as

\begin{equation}\label{equ:loss_sl}
	\mathcal{L}_{\text {sl}}=-\frac{1}K \sum_k \sum_c \bar{y}_{kc} \log \bar{z}_{kc} +\left(1-\bar{y}_{kc} \right)\log (1-\bar{z}_{kc})
\end{equation}
where $\bar{z}_{kc}$ denotes the probability of $\mathbf{s}_k$ over $c$-th category; $\bar{y}_{kc}$ equals 1 if $c$-th category exists in $\mathbf{s}_k$, otherwise 0.

Moreover, we present the additional contextual constraint to restrict point-wise predictions. The intuition behind is that for each $\mathbf{s}_k$, point-wise predictions should be limited to OCOC label $\textbf{l}_k$. Global max pooling is used to calculate the class-wise maximum probability $\bar{p}$ of $\mathbf{s}_k$. With $\bar{p_c}=\max_{i} p_{ic}$, $\bar{p}$ is used to imply the probability for the occurrence of each semantic category in $\mathbf{s}_k$. We also adopt BCE for loss calculation of global maximum probability:

\begin{equation}\label{equ:max_prob}
	\mathcal{L}_{\text {gmp}}=-\frac{1}K \sum_k \sum_c \bar{y}_{kc} \log \bar{p}_{kc} +\left(1-\bar{y}_{kc} \right)\log (1-\bar{p}_{kc})
\end{equation}
where $\bar{p}_{kc}$ denotes the prediction probability of $\mathbf{s}_k$ over category $c$. And $\bar{y}_{kc}$=1 if $c$ exists in $\mathbf{s}_k$, otherwise 0. $\mathcal{L}_{\text {sl}}$ and $\mathcal{L}_{\text {gmp}}$ jointly enhance learning the semantic representation of point clouds.

\begin{figure*}[t]
\centering
\includegraphics[width=1\linewidth]{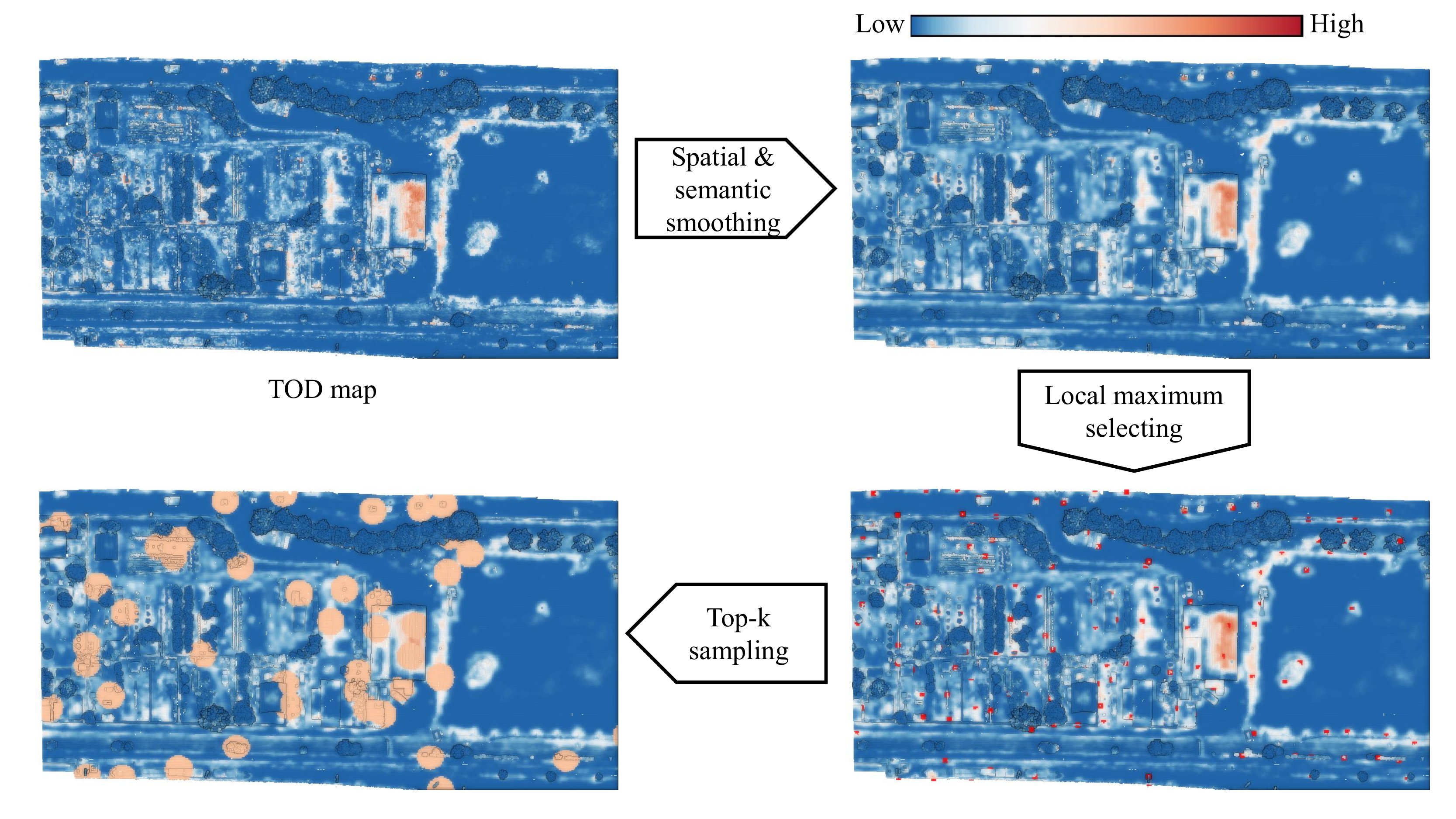}
\caption{Workflow of sub-clouds query. Given the TOD saliency map, we first conduct a spatial and semantic smoothing for outlier removal. Then, points with maximum value within a local region are extracted as candidates for sub-cloud center, and a top-ranked sampling is incorporated for sub-clouds selection.}
\label{fig:al}
\end{figure*}

\subsubsection{Context-aware pseudo labeling}
\label{sec:m_ws_pl}

As a simple yet efficient weakly supervised method, pseudo labels $l^{\text{pl}}$ are able to greatly alleviates annotation scarcity issue. For each sub-cloud $\mathbf{s}_k$, $l^{\text{pl}}$ of unlabeled points are generated based on predicted probabilities over the multi-class labels, shown as:
\begin{equation}\label{equ:5}
	 l^{\text{pl}}_i = \arg\max p_{ic}, c \in \mathbf{l}_k
\end{equation}
Generally, a class-wise prediction with high posterior probability is more likely to be correct. In the light of facts that fixing a generic threshold for pseudo label generation is not applicable to diverse datasets or scenes, we derive and soften pseudo labels for all unlabeled data $\mathbf{s}_{u}$ by associating them with different weights $w$ based on classification uncertainty. In this study, the Shannon entropy \citep{shannon} $H$ of the predicted probability for each semantic class is utilized to measure the uncertainty, and a larger value represents higher uncertainty, denoted as:

\begin{equation}\label{equ:entropy}
	H =-\sum_{c}^{C}p_{c}\log p_{c}
\end{equation}
Then, the weight $w_{i}$ of pseudo label for point $x_{i}$ is defined as:
\begin{equation}\label{equ:weight}
	 w_{i}=1-\frac{H_{i}}{\log C}
\end{equation}
$\log C$ normalizes $w_{i}$ to [0, 1] according to the principle of maximum entropy. The loss of pseudo labels is calculated by weighted cross-entropy:
\begin{equation}\label{equ:7}
	 \mathcal{L}^{\text{pl}}_{\text{seg}} = -\frac{1}{\left | \mathbf{s}_{u} \right |}\sum_{i}^{M_u} w_{i} \sum_{c}^{C} y^{\text{pl}}_{ic}\log p_{ic}
\end{equation}

\subsubsection{Loss summation}
\label{sec:m_ws_loss}

All the losses simultaneously engage in the backpropagation process for network training. The combined optimization is presented as follows:

\begin{equation}\label{equ:loss}
	 \Theta=\arg \min_{\Theta} (\mathcal{L}_{\text{seg}} + \mathcal{L}_{\text{sl}} + \mathcal{L}_{\text{gmp}} + \mathcal{L}^{\text{pl}}_{\text{seg}})
\end{equation}
Since we propose to perform the model training in an active learning manner, $\mathcal{L}^{\text{pl}}_{\text{seg}}$ is not considered during the first training cycle with an aim to avert unreliable pseudo labels.

\subsection{Deep batch active learning}
\label{sec:m_al}

To identify most informative samples and minimize labeling costs, we further propose to incorporate an active learning method. Batch-based sample querying strategy is adopted in our study, which secures an efficient training pattern with deep models. Instead of a greedy search to analyze all possible sub-clouds for retrieving most informative ones, point-wise loss is first estimated, guiding top-ranked sub-clouds extraction followed by corresponding OCOC annotation. The rational is built on two perceptions: first, since every single point could be chosen as the center to construct a sub-cloud, it is less practical to evaluate numerous potential sub-clouds from the whole point cloud for comparison and selection; Second, our task is aimed at point cloud semantic segmentation, thus, desirable sub-clouds are highly associated with those local regions containing points with high uncertainty. Under this situation, we develop a two-step query strategy, which imposes point-wise uncertainty to discover informative sub-clouds. The pipeline of the proposed active learning method is presented in Alg.~\ref{alg:al}.

\begin{algorithm}[b!]
\caption{Active Weakly Supervised Point Cloud Semantic Segmentation}\label{alg:al}
\renewcommand{\algorithmicrequire}{\textbf{Input:}}
\renewcommand{\algorithmicensure}{\textbf{Output:}}
\algnewcommand{\LeftComment}[1]{\Statex \(\triangleright\) #1}
\begin{algorithmic}

\Require Point clouds ${\mathbf{P}\in\mathbb{R}^{N\times D}}$,
\Ensure Predictions ${p\in\mathbb{R}^{N\times C}}$
\State \textit{\# OCOC initialization}
\State $\mathbf{S}=\{\mathbf{s}_1, \mathbf{s}_2, \dots, \mathbf{s}_K\}$
\Comment{Randomly sample $K$ sub-clouds}
\For{$\mathbf{s}$ in $\mathbf{S}$}
   \For{$c$ in $\bar{l}_s$}
       \State $x \leftarrow c$
       \Comment{Label one point belonging to $c$}
   \EndFor
\EndFor
\\
\State \textit{\# Active weakly supervised learning}
\For{each training cycle $\Lambda$}
    \State \textit{\# Model training}
    \Repeat
       \State Train model for one epoch $\Theta^{\{\Lambda\}}$:
       \State $\Theta^{\{\Lambda\}} = \Theta^{\{\Lambda\}} - \eta \nabla (\mathcal{L}_{seg} + \mathcal{L}_c + \mathcal{L}_{gp} + \mathcal{L}^{pl}_{seg})$;
    \Until convergence
    \\
    \State \textit{\# Sub-clouds query}
    \State $p^{\{\Lambda\}}=\Theta^{\{\Lambda\}}(\mathbf{P})$
    \Comment{Infer predictions}
    \State $D^{\{\Lambda\}}=\left\|p^{\{\Lambda\}}-p^{\{\Lambda-1\}}\right\|^2$
    \Comment{Calculate TOD}
    \State $\tilde{D}^{\{\Lambda\}}=f_w(D^{\{\Lambda\}}, w^p, w^s)$
    \Comment{Refine TOD}
    \State $\textbf{x}_s\in LocalMaximum(\tilde{D}^{\{\Lambda\}})$
    \Comment{Select seed points}
    \\
    \State \textit{\# Top-k sampling}
    \State $\textbf{S}^{\{\Lambda\}}$ = \O
    \For{$k \leftarrow 1$ to $K$}
        \State $\textbf{s}$ = sub-cloud$(\arg\max{\textbf{x}_s})$
        \State $\textbf{s} \gets$ TOD guided OCOC 
        \State $\textbf{S}^{\{\Lambda\}}$ = $\textbf{S}^{\{\Lambda\}} \cup \textbf{s}$
        \State $\textbf{x}_s = \textbf{x}_s \setminus x\in \textbf{s}$
        \Comment{Discard utilized seeds}
    \EndFor
    \State $\textbf{S} =\textbf{S} \cup \textbf{S}^{\{\Lambda\}}$
    \Comment{Expand labeled pool}
\EndFor

\end{algorithmic}
\end{algorithm}

\subsubsection{Temporal output discrepancy}
\label{sec:m_al_info}

Measuring the sample uncertainty plays a vital role in active learning, as it enables the learner to query the example which it is least certain about. Compared with commonly used strategies in classical active learning methods - uncertainty sampling (e.g., margin sampling~\citep{Scheffer2001} or entropy~\citep{shannon}), loss is regarded as a perfect indicator for uncertainty measurement in deep learning \citep{Yoo_2019_CVPR}. In order to estimate the prediction loss for unlabeled points, we adopt Temporal output discrepancy (TOD) \citep{Huang_2021_ICCV} to infer informative samples, which is theoretically associated with loss function, being efficient and flexible to implement. Given a sample $x$ and the neural network $f$, TOD $D_t^{\{T\}}(x)$ is defined as

\begin{equation}\label{equ:tod}
	 D_t^{\{T\}}(x) \stackrel{\text { def }}{=}\left\|f\left(x ; w_{t+T}\right)-f\left(x ; w_t\right)\right\| 
\end{equation}
where $f(x; w_t)$ stands for the output of model $f$ over $x$ with parameters $w_t$ at the $t$-th training step. Thus, TOD describes the temporal prediction distance between different training steps.

TOD is first associated with loss of the one-step situation, which is provided in \citet{Huang_2021_ICCV} such that

\begin{equation}\label{equ:tod_1step}
	 D_t^{\{1\}}(x) \leq \eta \sqrt{2 \mathcal{L}_t(x)}\left\|\nabla_w f\left(x ; w_t\right)\right\|^2
\end{equation}
where $\eta$, $\mathcal{L}_t(x)$ and $\nabla_w f$ are learning rate, sample loss and gradient, respectively. Note that here $\mathcal{L}_t(x)=\frac{1}{2}\left(y-f\left(x ; w_t\right)\right)^2$, and similar results can be also observed with cross-entropy loss. With Equ.~\ref{equ:tod_1step}, we can further deduce

\begin{equation}\label{equ:tod_tstep}
	 D_t^{\{T\}}(x) \leq \sqrt{2} \eta \sum_{\tau=t}^{t+T-1}\left(\sqrt{\mathcal{L}_\tau(x)}\left\|\nabla_w f\left(x; w_\tau\right)\right\|^2\right)
\end{equation}
Then, from the proof that $\nabla_w f$ can be deemed approximately as a constant $\mu$, denoted as $\left\|\nabla_w f\right\|^2 \leq \mu$ , $D_t^{\{T\}}(x)$ is finally deduced with Cauchy–Schwarz inequality:

\begin{align}\label{equ:tod_loss}
\begin{split}
   D_t^{\{T\}}(x) & \leq \sqrt{2} \eta \mu \sum_{\tau=t}^{t+T-1} \sqrt{\mathcal{L}_\tau(x)} \\ & \leq \sqrt{2 T} \eta \mu \sqrt{\sum_{\tau=t}^{t+T-1} \mathcal{L}_\tau(x)} 
\end{split}
\end{align}
$D_t^{\{T\}}$ represents a lower bound of $\mathcal{L}$ during $T$ training steps, which is an effective indicator to approximate $\mathcal{L}$. In this study, we use $D_t^{\{T\}}$ for point-wise loss estimation. We directly calculate squared sum of prediction discrepancy of each point $x_i$ for training cycle $\Lambda$ as TOD:
\begin{equation}\label{equ:tod_point}
	 D^{\{\Lambda\}}(x_i) = \left\|p^{\{\Lambda\}}_{x_i}-p^{\{\Lambda-1\}}_{x_i}\right\|^2
\end{equation}
Note that, in the first training cycle, TOD is computed based on the prediction discrepancy between the initialized model and the trained model.

\subsubsection{Extraction of informative sub-clouds }
\label{sec:m_al_sc}

We explore informative sub-clouds with help of the derived TOD saliency map. The pipeline of sub-clouds selection is presented in Fig.~\ref{fig:al}. Due to the very high point density of LiDAR data, sub-clouds which are centered on adjacent points are almost not distinguished from each other. To realize an efficient query, we first downsample the TOD map with the resolution of one fifth of sub-cloud radius size. Then, a local smoothing strategy is proposed to eliminate outliers and preserve TOD consistency, which contributes to locating reliable uncertain areas. A $k$-nn graph $\{x_1, x_2, ..., x_k\}$ is constructed for each point $x$, with distance $d_i = \|x-x_i\|^2, i\in\{1, 2, ..., k\}$. The corresponding weight of edge is formulated with respect to the cosine similarity of predicted probabilities $w^p$ and spatial distance $w^s$:
\begin{equation}\label{equ:knn}
	 w_i = w^p_i \cdot w^s_i = \frac{p_{x} \cdot p_{x_{i}}}{\|p_{x}\|\cdot\|p_{x_{i}}\|}\cdot (1-\frac{d_i}{\max{d_i}})^2, i\in\{1, 2, ..., k\}
\end{equation}
Refined TOD is calculated as
\begin{equation}\label{equ:tod_refine}
	 \tilde{D}(x)=\frac{\sum_{i}^{k} w_i D(x_i) }{\sum_{i}^{k} w_i}
\end{equation}

Apart from concerned information measure - TOD, diversity among a batch of samples is also of great importance. Due to the lack of a preexisting sub-cloud pool for distribution modeling and analysis, we explicitly advocate the spatial diversity and propose a top-k sampling method for querying the batch of sub-clouds. Conventional top-k sampling algorithms often lead to overconcentration in the nearby region, which triggers redundancy issue. Inspired from the first law of geography \citep{Tobler} that ``near things are more related than distant things'', we sample sub-clouds with spatial diversity using a local maximum filtering strategy. Points with local maximum $\tilde{D}$ are selected as the initial seeds to represent their localized areas with the same size as considered sub-clouds. Then, points with top-k maximum TOD is chosen as the center point to extract the sub-clouds. During the batch query, once a sub-cloud is extracted, the contained seed points are discarded to reduce the redundancy.

\begin{figure}[!b]
\centering
\includegraphics[width=1\linewidth]{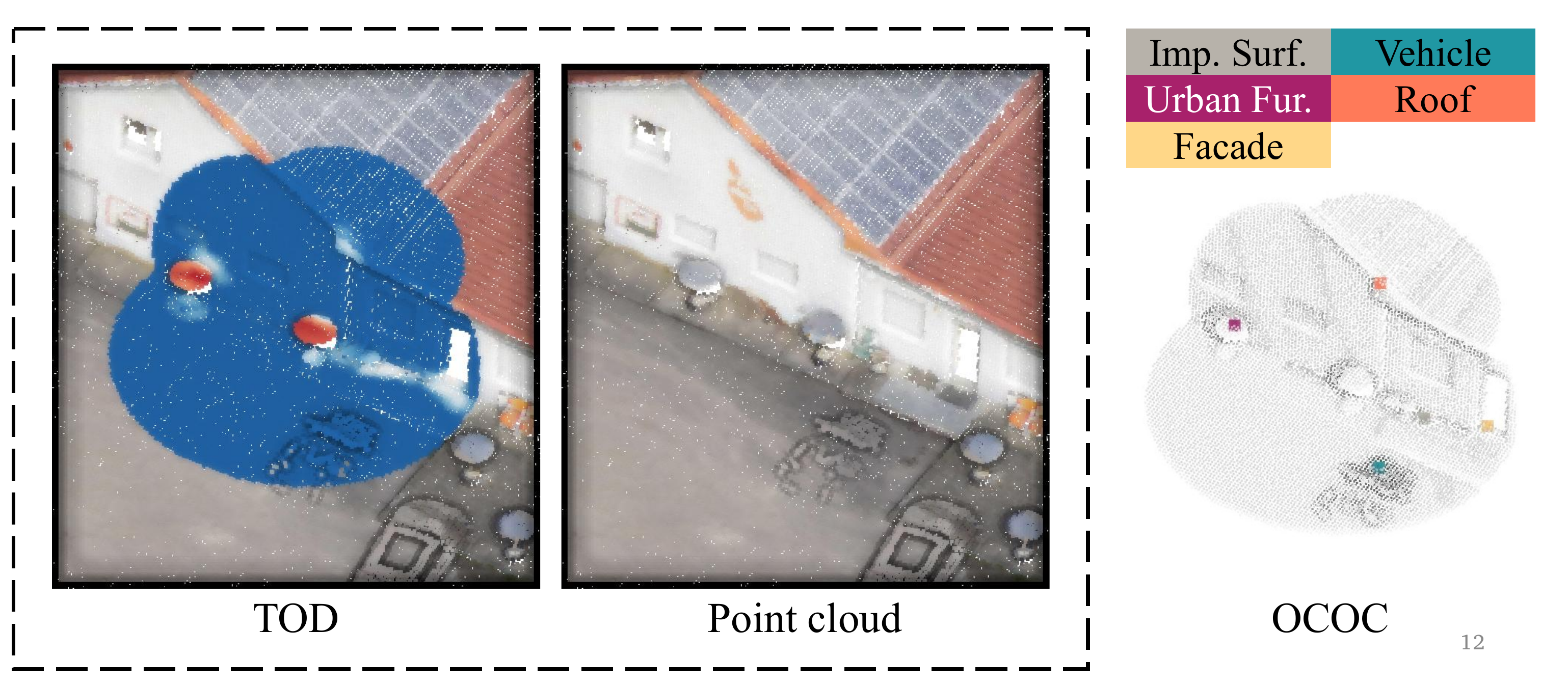}
\caption{TOD guided OCOC annotation. Given a sub-cloud with the corresponding TOD saliency map, we believe the operator is capable and inclined to label points with distinctly high values.}
\label{fig:annotating}
\end{figure}

\subsubsection{TOD guided OCOC labeling}
\label{sec:m_al_label}

For weak supervision task, compared to conducting OCOC randomly within a sub-cloud, we argue that labeling with effective guidance is more conducive to attaining  desirable ones. Since we evaluate our method on open-access benchmarks, we introduce a virtual labeling design for active learning, which is illustrated in Fig.~\ref{fig:annotating}. Given a sub-cloud $\mathbf{s}_k$ and corresponding TOD saliency map $D(\mathbf{s}_k)$, we assume that the operators are able and inclined to annotate points with high $D$. Moreover, according to visual perception, human operators tend to annotate points in salient areas. To this end, we further design a decay mechanism to avoid boundary regions.

First, we believe the annotators often circumvent points close to object edges, thus a sub-cloud boundary index $b$ is formulated as $b=1-\exp(-r(r-d))$ to penalize boundary points, where $r$ and $d$ denote the sub-cloud radius size and plane distance to the center point, respectively. Additionally, to avoid semantic boundary, for point $x$, a pair-wise local semantic homogeneity $h$ is calculated from a k-nn graph $\{x_1, x_2, ..., x_k\}$, formulated as $h_i=\text{boolean}(l_x==l_{x_i})$. Then, we update TOD value within each sub-cloud as 
\begin{equation}\label{equ:tod_sc}
     D^{\prime}(x) = \frac{\sum_{i}^{k} D(x_i) h_i}{k} \cdot b_x
\end{equation}
Points with maximum $D^{\prime}$ at each category are annotated, which simulates the human labeling process under TOD guidance.

\section{Experiment}
\label{sec:exp}

\subsection{Dataset description}
\label{sec:exp_data}
Three high-density LiDAR point cloud benchmarks collected from multiple ground/aerial platforms were chosen for evaluation and analysis: an ALS dataset Hessigheim 3D (H3D) \citep{H3D}, a MLS dataset Paris-Lille-3D (Paris3D) \citep{Paris3D}, and a TLS dataset Semantic3D \citep{Semantic3D}. 

\paragraph{Hessigheim 3D}
The dataset comprises a high-density LiDAR point cloud of approximately 800 points/m² enriched with an RGB image of 2-3 cm GSD, acquired from a Riegl VUX-1LR scanner and two oblique-looking Sony Alpha 6000 cameras mounted on a RIEGL Ricopter platform. The study area of interest is Hessigheim, Germany. The entire study area is divided into three connected sections for training, validation, and testing. The training and validation sets are used in this study, for which the number of points was approximately 59.4 million and 14.5 million, respectively. Eleven semantic categories are predefined, including low vegetation, impervious surfaces, vehicles, urban furniture, roofs, facades, shrubs, trees, soil/gravel, vertical surfaces, and chimneys. The format of the utilized features was \{X, Y, Z, R, G, B\}.

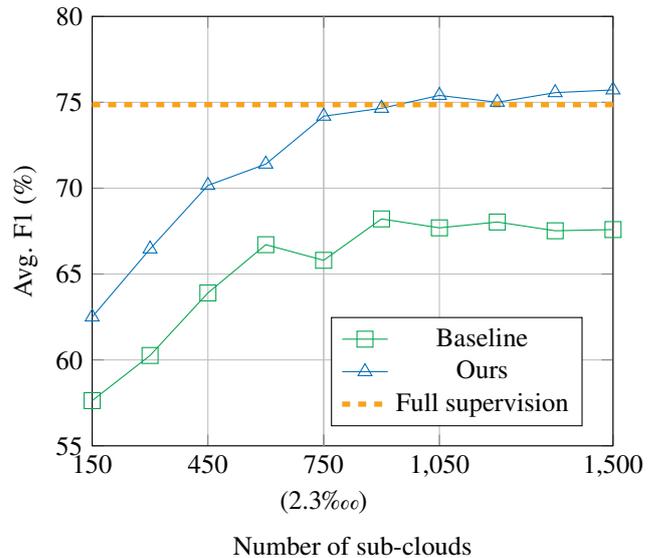
\begin{figure*}[b]
\centering
\begin{subfigure}[b]{0.49\textwidth}
    \centering
    \begin{tikzpicture}
    \begin{axis}[
    grid=both,
    xmin=150, xmax=1500,
    ymin=70, ymax=90,
    xtick={150, 450, 750, 1050, 1500},
    extra x ticks ={750},
    extra x tick labels={\begin{tabular}[c]{@{}c@{}} \\ (2.3\textpertenthousand{})\end{tabular} },
    xlabel=Number of sub-clouds, 
    ylabel=OA (\%), 
    legend style={at={(0.7,0.3)},anchor=north}
    ]
    \addplot[mark=square, mark size=3pt, Green] plot coordinates { 
     (150,75.69)
     (300,73.25)
     (450,77.21)
     (600,80.05)
     (750,78.94)
     (900,81.83)
     (1050,82.54)
     (1200,82.74)
     (1350,83.54)
     (1500,83.63)
    };
    \addlegendentry{Baseline}
 
    \addplot[mark=triangle, mark size=3pt, NavyBlue] plot coordinates {
     (150,80.10)
     (300,82.72)
     (450,84.60)
     (600,85.62)
     (750,86.46)
     (900,86.83)
     (1050,87.02)
     (1200,87.25)
     (1350,87.61)
     (1500,87.74)
    };
    \addlegendentry{Ours}

    \addplot[dashed, YellowOrange,line width=2pt, domain=150:1500] {87.18};
    \addlegendentry{Full supervision}
    \end{axis}

    \end{tikzpicture}
\end{subfigure}
\hfill
\begin{subfigure}[b]{0.49\textwidth}
    \centering

    \begin{tikzpicture} 
    \begin{axis}[
    grid=both,
    ymin=55, ymax=80,
    xmin=150, xmax=1500,
    xtick={150, 450, 750, 1050, 1500},
    extra x ticks ={750},
    extra x tick labels={\begin{tabular}[c]{@{}c@{}} \\ (2.3\textpertenthousand{})\end{tabular} },
    xlabel=Number of sub-clouds, 
    ylabel=Avg. F1 (\%),
    legend style={at={(0.7,0.3)},anchor=north}
    ]
    \addplot[mark=square, mark size=3pt, Green] plot coordinates 
    { 
     (150,57.63)
     (300,60.26)
     (450,63.90)
     (600,66.71)
     (750,65.80)
     (900,68.21)
     (1050,67.69)
     (1200,68.03)
     (1350,67.52)
     (1500,67.59)
    };
    \addlegendentry{Baseline}
 
    \addplot[mark=triangle, mark size=3pt, NavyBlue] plot coordinates {
     (150,62.50)
     (300,66.46)
     (450,70.16)
     (600,71.40)
     (750,74.19)
     (900,74.65)
     (1050,75.40)
     (1200,75.00)
     (1350,75.56)
     (1500,75.72)
    };
    \addlegendentry{Ours}

    \addplot[dashed, YellowOrange, line width=2pt, domain=150:1500] {74.86};
    \addlegendentry{Full supervision}
    \end{axis}
    \end{tikzpicture}
\end{subfigure}

\caption{The impact of the increased number of sub-cloud labels on OA and Avg. F1 for H3D dataset.}
\label{fig:h3d_line}
\end{figure*}

\paragraph{Paris-Lille-3D}
The dataset was acquired acquired with a MLS system, which was equipped with a GPS, an inertial measurement unit, and a Velodyne HDL-32E LiDAR. The point clouds feature high density with between 1,000 and 2,000 points/m² on the ground. The dataset consists of three parts in France, namely two parts in the agglomeration of Lille and one in Paris, covering a 2-km long street environment with 143.1 million points. For efficient evaluation of our proposed method, we extract \textit{Lille2} file from the training set for validation. Nine main classes are considered for point cloud semantic segmentation task, including ground, building, pole, bollard, trash can, barrier, pedestrian, car, and natural. Unclassified points without predefined labels in the dataset were excluded from the training process. The format of the utilized features was \{X, Y, Z\}.

\paragraph{Semantic3D}
The dataset was captured from surveying-grade TLS system, consisting of total 4 billions 3D points. Moreover, colorization was performed by deploying a high resolution cubemap generated from camera images. The study scenes comprise typical Central European architecture in urban and rural areas. The original training set includes 15 dense point clouds. Still, four files were excluded from the training set for testing, including \textit{bildstein5}, \textit{domfountain3}, \textit{untermaederbrunnen3}, and \textit{sg27\_9}. Eight classes are manually annotated, covering man-made terrain, natural terrain, high vegetation, low vegetation, buildings, hard scape, scanning artefacts, and cars. We discarded unclassified points predefined in the dataset before training. The format of the utilized features was \{X, Y, Z, R, G, B\}.

\subsection{Implementation}
\label{sec:exp_imp}

We illustrate some configuration details here. First, considering the high density of raw data, we subsampled the data in advance to improve computational efficiency while preserving point cloud structural details. During training, the grid size was set to 0.1m for H3D dataset, 0.08m for other two. In the inference process, predictions of raw test data are obtained from nearest-neighbor interpolation. Moreover, for Semantic3D dataset, considering the extremely high point density, uniform downsampling with a spacing of 0.01 m was performed in advance on the testing set for evaluation, which keeps the consistency with the default benchmark setting. For mini-batch generation, spherical shape sub-cloud were considered, and the radius size for H3D, Paris3D, and Semantic3D were 5m, 4m, and 4m, respectively. During training, as the number of points in each sub-cloud varies, the batch size is not fixed, and the upper bound of the total number of points at each training step is set to 12,000 in this study. During the test process, overlapping sub-clouds were considered, and each point was tested approximately three times. We basically took over the default parameters of the KPConv segmentation network, adopting an Adam optimizer with an initial learning rate of $10^{-3}$. All models were implemented within the PyTorch framework. 

\subsection{Weak label configuration}
\label{sec:exp_wl}

For each training cycle of active learning, based on the scale of the dataset, the number of newly added sub-clouds amounts to 150, 300, and 220 for H3D, Paris3D, and Semantic3D datasets, respectively. The locations of initial sub-clouds and corresponding weak labels are determined with random selection. We define the  scarcity level for weak labels as the percentage of point-level annotations to subsampled point clouds during training.   

\subsection{Evaluation metrics}
\label{sec:exp_val}

We used the overall accuracy (OA) and F1 scores to evaluate the performance of proposed method. OA stands for the percentage of points that are predicted with correct labels, while the F1 score denotes the harmonic mean of the precision and recall, expressed as:
\begin{equation}\label{equ:validation}
\begin{aligned}
& precision = \frac{{tp}}{{tp+ fp}},\\
& recall = \frac{{tp}}{{tp + fn}},\\
& F1 = 2 \times \frac{{precision \times recall}}{{precision + recall}},\\
\end{aligned}
\end{equation}
where $tp$, $fp$, and $fn$ are true positives, false positives, and false negatives, respectively.

\begin{table*}[!b]
\caption{Comparison of full and weak supervisions on H3D dataset}
\label{tab:h3d}
\centering
\small
\setlength\tabcolsep{2pt}%
\begin{tabularx}{\textwidth}{ccXXXXXXXXXXXXX}
\hline
\multirow{3}{*}{Setting} & \multirow{3}{*}{Method} & \multicolumn{11}{c}{F1 Score} & \multirow{3}{*}{Avg. F1} & \multirow{3}{*}{OA} \\ 
\cline{3-13} &  & {\begin{tabular}[c]{@{}c@{}}Low \\ veg.\end{tabular}} & {\begin{tabular}[c]{@{}c@{}} Imp.\\ Surf.\end{tabular}} & Vehicle & {\begin{tabular}[c]{@{}c@{}} Urban\\ Fur. \end{tabular}}& Roof & Facade & Shrub & Tree & Soil & {\begin{tabular}[c]{@{}c@{}}Ver.\\Surf. \end{tabular}} & Chimney &    &  \\ \hline
\multirow{2}{*}{Full Sup.}                                      
& RandLA-Net  & 88.97  & 90.50  & 47.77  & 63.11   & 96.81  & 79.00  & 62.16  & 95.06 & 45.49 & 74.52  & 87.17   & 75.53  & 88.09  \\
& KPConv  & 88.64  & 87.73  & 80.81  & 63.47  & 94.71  & 78.12   & 61.71  & 95.57  & 33.21  & 71.08  & 68.43  & 74.86  & 87.18  \\ \hline
\multirow{2}{*}{\begin{tabular}[c]{@{}c@{}}Weak Sup. \\ (2,873 sc) \end{tabular}} 
& MPRM   & 85.37  & 81.81  & 41.42  & 33.95  & 91.89  & 68.01  & 41.91  & 89.78  & 47.53  & 48.83  & 0.00    & 57.32   & 81.76   \\
& Weak-ALS  & 76.13  & 80.00  & 11.51  & 39.15  & 88.52  & 75.06  & 36.61  & 87.78  & 3.09  & 47.26  & 0.00  & 49.56  & 75.71  \\ \hline
\multirow{2}{*}{\begin{tabular}[c]{@{}c@{}}Weak Sup. \\  (750 sc) \end{tabular}}  
& Baseline  & 79.18  & 78.81  & 45.55  & 55.38  & 90.06  & 72.62  & 53.28  & 93.53  & 36.00  & 57.20  & 62.21  & 65.80  & 78.94  \\
& Ours  & 87.38  & 86.72  & 68.84  & 60.81  & 94.30  & 78.58  & 60.00  & 95.31  & 17.96  & 81.18  & 84.96  & 74.19   & 86.46  \\ \hline
\end{tabularx}
\end{table*}

\section{Results and discussion}
\label{sec:res}

\subsection{Semantic segmentation results}
\label{sec:res_res}

\subsubsection{H3D dataset}
\label{sec:res_res_h3d}

\begin{figure}[t]
\centering
\begin{subfigure}[b]{1.0\linewidth}
    \centering
    \includegraphics[width=1\linewidth]{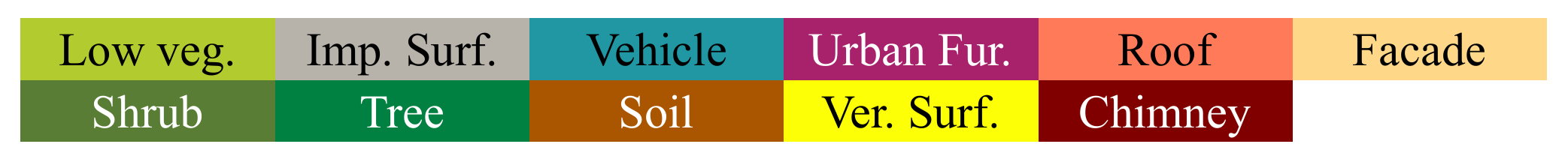}
\end{subfigure}
\hfill
\includegraphics[width=1\linewidth]{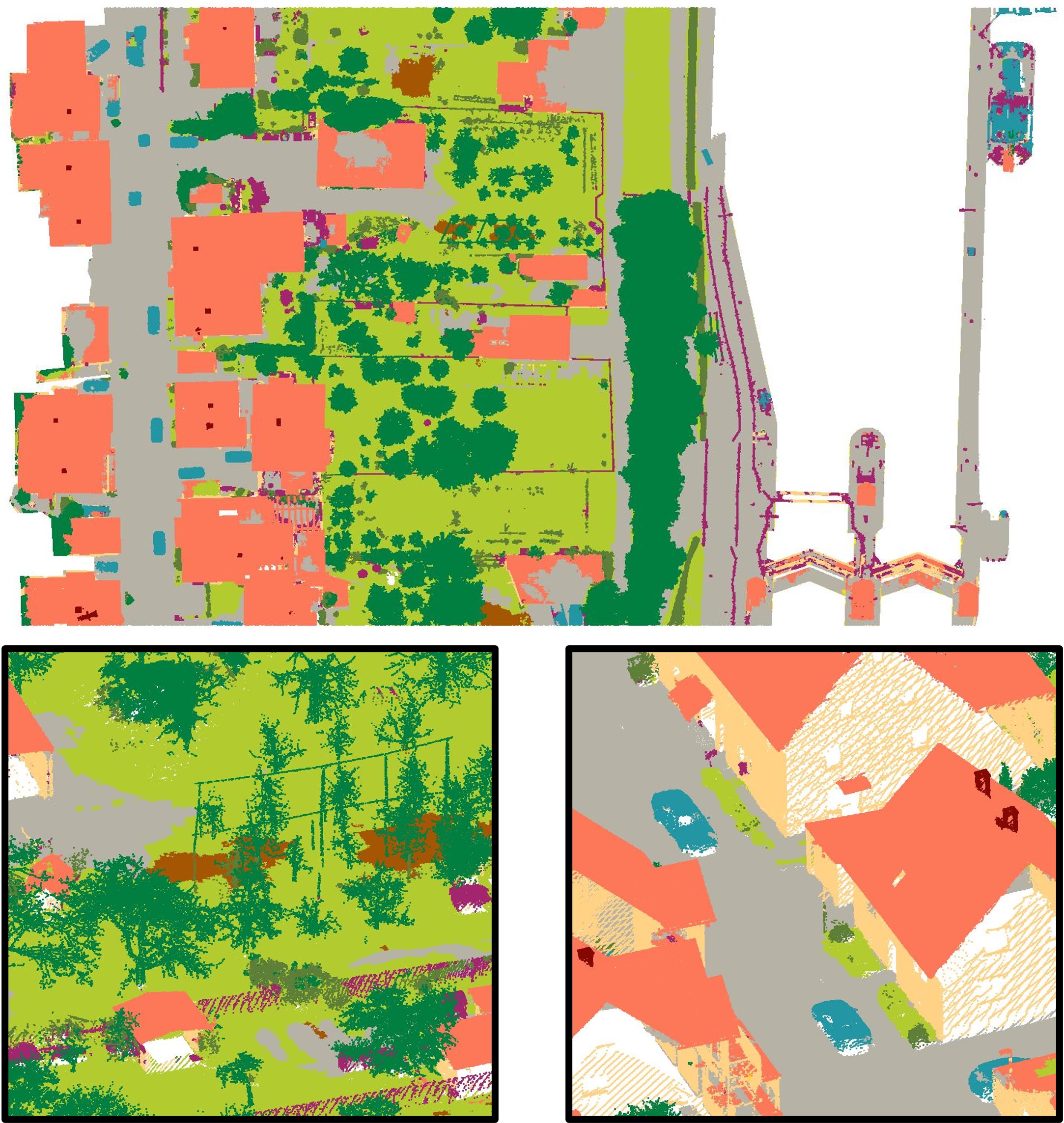}
\caption{H3D dataset classification map using 750 sub-clouds.}
\label{fig:h3d_res}
\end{figure}

We first show our weak supervision results with active learning. The baseline method stands for the situation that only $\mathcal{L}_{\text{seg}}$ is adopted for back propagation, and sub-clouds and corresponding OCOC are randomly generated during each training cycle. Full supervision result of KPConv is also listed for comparison. The accuracy variation with respect to increasing OCOC is shown in Fig.~\ref{fig:h3d_line}. From the results, it is obvious that the accuracy is improved along with increased number of the sub-clouds  for both the baseline and our methods, which is consistent with the general perception of active learning. Compared with the baseline, the two evaluation indices have significantly increased with our method. When the number of sub-clouds increases to 750, which contains 2.3\textpertenthousand{} of total labels, the performance of our method is close to full supervision ones. Then, results tend to be robust with subsequent label expansion. After 10 training cycles, even better average F1 score is achieved compared to the full supervision scheme. In light of the trade-off between the annotated data amount and model performance, the promising result shows exceptional competence in reducing the labeling costs. Fig.~\ref{fig:h3d_res} presents the classification result using 750 sub-clouds. Most of points are classified correctly, which apparently means the semantic boundary is preserved as well. Additionally, from the local region visualization, small objects such as cars and chimney can be clearly recognized. This implies that our method addresses the issue of semantic discontinuity of sparse labels, exporting readily recognizable/parsed object classes. 

We compare our method with previously published works using both fully-and weakly supervised learning methods, which are all based on deep models. First, we introduce two deep networks under full supervision scheme, which are chosen owing to their proven performance. RandLA-Net~\citep{hu2020randla} proposed an efficient point cloud semantic segmentation network with randomly downsampling. Local spatial encoding and attentive pooling were used for effective feature aggregation. KPConv is used as the backbone network in this study, and we also assess its performance under full supervision. Regarding weakly supervised point cloud learning, we choose two methods based on scene-level labels for implementing the experiments on the utilized datasets. MPRM~\citep{wei2020multi} proposed to localize point-wise semantic labels based on multi-path point class activation maps. \citet{LIN202279} tried to enhance MPRM by introducing an overlapp region loss and contrastive constraint for pseudo labels, which is referred to as Weak-ALS in this study. For a fair comparison, we use the same radius size as that in our method to sample sub-clouds. In these two weak supervision methods, sub-clouds are obtained by uniform sampling, leading to a much larger annotation load. 

The quantitative comparison results are presented in Table~\ref{tab:h3d}. Using 750 sub-clouds, compared with the baseline, the OA and F1 scores of all categories are considerably improved using our method, especially for marginal categories such as vehicle and chimney. In contrast, other two weakly supervised methods show unsatisfactory results though more sub-clouds are used, and some categories are completely misclassified. This indicates the shortcoming of methods which barely rely on leveraging scene-level labels. Without point-level annotations, the deep network fails to conduct reasonable point-wise inference through decoding layers. Thus, it is difficult to identify accurate semantic boundary, and marginal categories are prone to be ignored. As for methods under full supervision, there is only a small performance gap between our method and two full supervision ones in terms of overall accuracy. This demonstrates that our method effectively counteracts the absence of semantic boundary information imposed by sparse annotations.

\begin{figure*}[t]
\centering
\begin{subfigure}[b]{0.49\textwidth}
    \centering
    
    \begin{tikzpicture}
    \begin{axis}[
    grid=both,
    ymin=94, ymax=100,
    xmin=300, xmax=3000,
    xtick={300, 900, 1500, 2100, 3000},
    extra x ticks ={1500},
    extra x tick labels={\begin{tabular}[c]{@{}c@{}} \\ (2.1\textpertenthousand{})\end{tabular} },
    xlabel=Number of sub-clouds, 
    ylabel=OA (\%), 
    legend style={at={(0.7,0.3)},anchor=north}
    ]
    \addplot[mark=square, mark size=3pt, Green] plot coordinates {
     (300,94.47)
     (600,95.29)
     (900,96.69)
     (1200,96.36)
     (1500,96.92)
     (1800,97.06)
     (2100,97.21)
     (2400,97.07)
     (2700,97.42)
     (3000,97.14)
    };
    \addlegendentry{Baseline}
 
    \addplot[mark=triangle, mark size=3pt, NavyBlue] plot coordinates {
     (300,96.77)
     (600,97.53)
     (900,97.96)
     (1200,98.12)
     (1500,98.57)
     (1800,98.51)
     (2100,98.70)
     (2400,98.71)
     (2700,98.56)
     (3000,98.59)
    };
    \addlegendentry{Ours}

    \addplot[dashed, YellowOrange,line width=2pt, domain=300:3000] {98.31};
    \addlegendentry{Full supervision}
    \end{axis}

    \end{tikzpicture}
\end{subfigure}
\hfill
\begin{subfigure}[b]{0.49\textwidth}
    \centering

    \begin{tikzpicture} 
    \begin{axis}[
    grid=both,
    ymin=60, ymax=95,
    xmin=300, xmax=3000,
    xtick={300, 900, 1500, 2100, 3000},
    extra x ticks ={1500},
    extra x tick labels={\begin{tabular}[c]{@{}c@{}} \\ (2.1\textpertenthousand{})\end{tabular} },
    xlabel=Number of sub-clouds, 
    ylabel=Avg. F1 (\%), 
    legend style={at={(0.72,0.28)},anchor=north}
    ]
    \addplot[mark=square, mark size=3pt, Green] plot coordinates { 
     (300,61.52)
     (600,63.82)
     (900,70.74)
     (1200,70.60)
     (1500,69.9)
     (1800,72.33)
     (2100,72.93)
     (2400,75.57)
     (2700,76.00)
     (3000,75.35)
    };
    \addlegendentry{Baseline}
 
    \addplot[mark=triangle, mark size=3pt, NavyBlue] plot coordinates {
     (300,70.50)
     (600,80.54)
     (800,86.07)
     (1200,87.26)
     (1500,90.30)
     (1800,89.16)
     (2100,89.91)
     (2400,90.11)
     (2700,89.12)
     (3000,89.85)
    };
    \addlegendentry{Ours}

    \addplot[dashed, YellowOrange,line width=2pt, domain=300:3000] {88.32};
    \addlegendentry{Full supervision}
    \end{axis}
    \end{tikzpicture}
\end{subfigure}

\caption{The impact of the increased number of sub-cloud labels on OA and Avg. F1 for Paris3D dataset.}
\label{fig:paris_line}
\end{figure*}
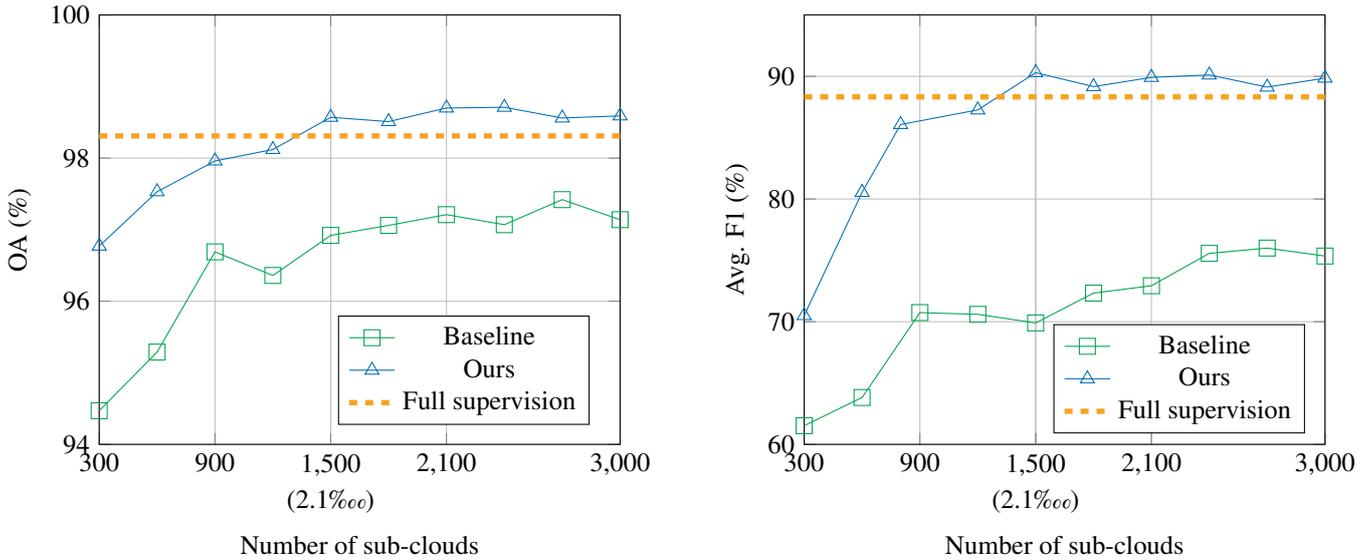

\begin{figure*}[!t]
\centering
\begin{subfigure}[b]{1.0\linewidth}
    \centering
    \includegraphics[width=0.8\linewidth]{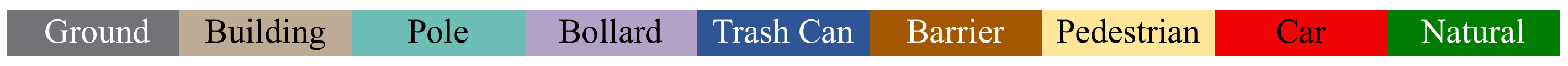}
\end{subfigure}
\hfill
\includegraphics[width=1\linewidth]{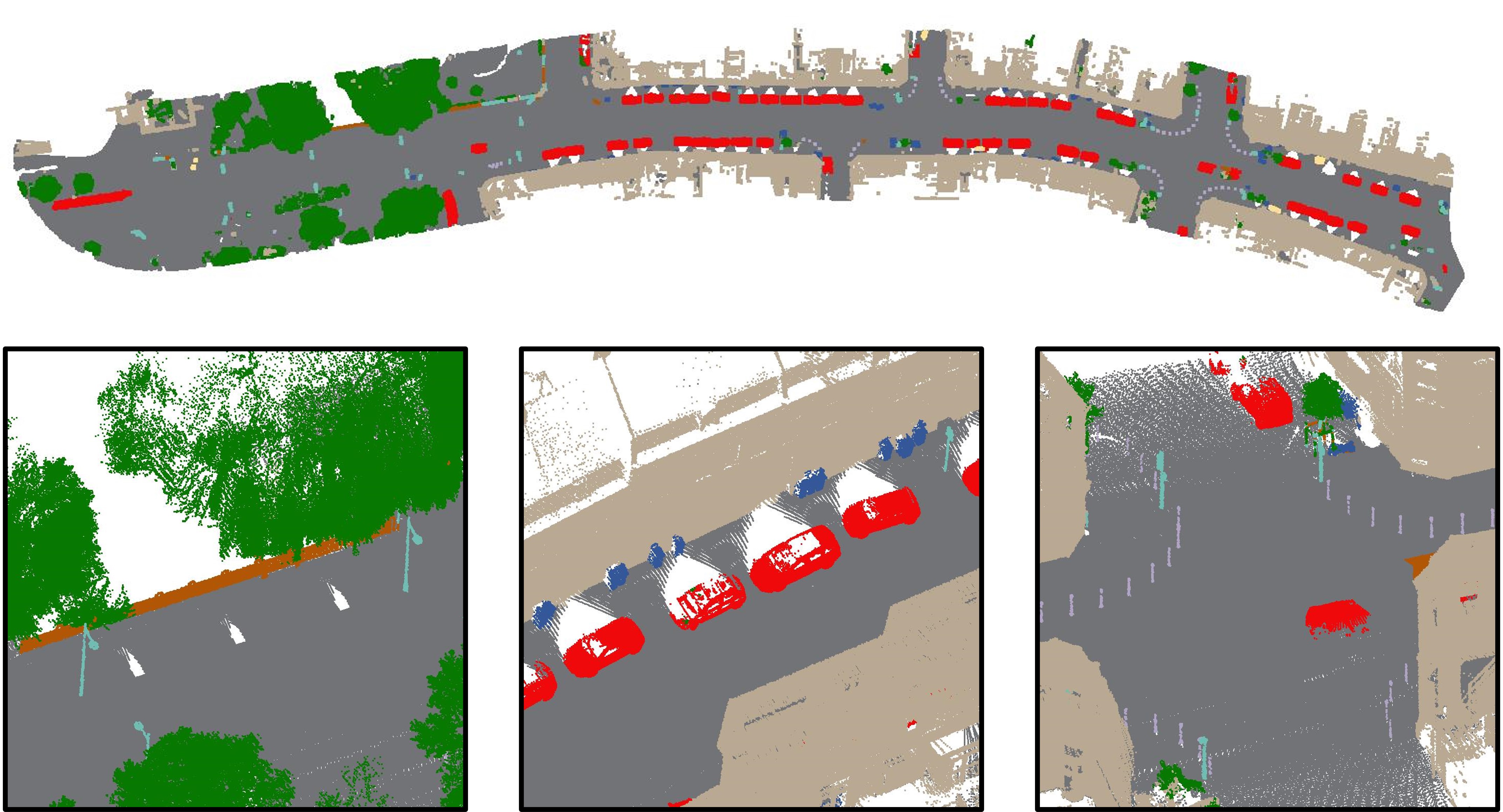}
\caption{Paris3D dataset classification map using 1,500 sub-clouds.}
\label{fig:paris_res}
\end{figure*}

\begin{table*}[t]
\caption{Comparison of full and weak supervisions for \textit{Lille2} data of Paris3D dataset}
\label{tab:paris}
\centering
\small
\setlength\tabcolsep{2pt}%
\begin{tabularx}{\textwidth}{ccXXXXXXXXXXX}
\hline
\multirow{3}{*}{Setting}  & \multirow{3}{*}{Method} & \multicolumn{9}{c}{F1 Score}  &  \multirow{3}{*}{Avg. F1} & \multirow{3}{*}{OA} \\ 
\cline{3-11}  &  &  Ground & Building & Pole & Bollard & {\begin{tabular}[c]{@{}c@{}}Trash \\ Can \end{tabular}}  & Barrier  & Pedestrian  & Car  & Natural  &  &  \\ \hline
\multirow{2}{*}{Full Sup.}                         
& RandLA-Net   & 99.26  & 99.15  & 84.97  & 88.45  & 81.92  & 41.92  & 95.12  & 98.79  & 87.39  & 86.33  & 98.34 \\
& KPConv  & 99.12  & 99.02  & 76.55  & 87.46  & 90.72  & 58.55  & 98.07  & 99.46  & 85.90  & 88.32  & 98.31  \\ \hline
\multirow{2}{*}{\begin{tabular}[c]{@{}c@{}}Weak Sup.  \\  (7,672 sc) \end{tabular}} 
& MPRM   & 92.78  & 94.33  & 64.35  & 2.42  & 65.28  & 37.32  & 52.72  & 69.00  & 88.53  & 62.98  & 90.80  \\
& Weak-ALS  & 98.00  & 97.02  & 63.35  & 19.12  & 63.62  & 42.45  & 28.24  & 94.37  & 79.65  & 65.09  & 96.02  \\ \hline
\multirow{2}{*}{\begin{tabular}[c]{@{}c@{}}Weak Sup. \\  (1,500 sc) \end{tabular}}  
& Baseline  & 98.97  & 97.45  & 62.19  & 38.48  & 74.58  & 31.14  & 46.76  & 98.30  & 81.23  & 69.90  & 96.92  \\
& Ours  & 99.28  & 99.26  & 85.21  & 90.83  & 90.10  & 63.98  & 97.03  & 98.35  & 88.33  & 90.30  & 98.57  \\ \hline
\end{tabularx}
\end{table*}

\begin{table*}[t]
\caption{Comparison of full and weak supervisions for testing set of Paris3D dataset}
\label{tab:paris_test}
\centering
\small
\begin{tabularx}{\textwidth}{ccXXXXXXXXXX}
\hline
\multirow{3}{*}{Setting}                                        
& \multirow{3}{*}{Method} & \multicolumn{9}{c}{IoU}  &  \multirow{3}{*}{Avg. IoU}\\ 
\cline{3-11}  & &  Ground & Building & Pole & Bollard & {\begin{tabular}[c]{@{}c@{}}Trash \\ Can \end{tabular}}  & Barrier  & Pedestrian  & Car  & Natural   \\ \hline
\multirow{2}{*}{Full Sup.}                         
& RandLA-Net   &99.5	&97.0	&71.0	&86.7	&50.5	&65.5	&49.1	&95.3	&91.7   &78.5 \\
& KPConv  &99.5	&94.0	&71.3	&83.1	&78.7	&47.7	&78.2	&94.4	&91.4   &82.0 \\ \hline
\multirow{2}{*}{\begin{tabular}[c]{@{}c@{}}Weak Sup. \\ \end{tabular}}  
& Ours (1,600 sc)  &99.4	&94.9	&53.6	&82.7	&59.5	&48.9	&60.9	&93.0	&88.8   &75.7  \\ 
& Ours (3,200 sc)  &99.4	&95.7	&59.7	&71.9	&70.6	&57.9	&66.7	&93.0	&90.8   &78.4  \\ \hline
\end{tabularx}
\end{table*}

\begin{figure*}[t]
\centering
\begin{subfigure}[b]{0.49\textwidth}
    \centering
    \resizebox{1.0\textwidth}{!}{
    \begin{tikzpicture}
    \begin{axis}[
    grid=both,
    ymin=80, ymax=100,
    xmin=220, xmax=2200,
    xtick={220, 440, 880, 1320, 1760, 2200},
    extra x ticks ={1320},
    extra x tick labels={\begin{tabular}[c]{@{}c@{}} \\ (2.0\textpertenthousand{})\end{tabular} },
    xlabel=Number of sub-clouds, 
    ylabel=OA (\%), 
    legend style={at={(0.7,0.3)},anchor=north}
    ]
    \addplot[mark=square, mark size=3pt, Green] plot coordinates { 
     (220,85.30)
     (440,88.31)
     (660,87.40)
     (880,90.03)
     (1100,90.31)
     (1320,90.56)
     (1540,89.01)
     (1760,91.90)
     (1980,91.77)
     (2200,92.25)
    };
    \addlegendentry{Baseline}
 
    \addplot[mark=triangle, mark size=3pt, NavyBlue] plot coordinates {
     (220,89.18)
     (440,93.21)
     (660,93.45)
     (880,94.06)
     (1100,94.17)
     (1320,94.61)
     (1540,94.31)
     (1760,94.20)
     (1980,94.55)
     (2200,94.96)
    };
    \addlegendentry{Ours}

    \addplot[dashed, YellowOrange, line width=2pt, domain=220:2200] {95.46};
    \addlegendentry{Full supervision}
    \end{axis}
    \end{tikzpicture}
    }
\end{subfigure}
\hfill
\begin{subfigure}[b]{0.49\textwidth}
    \centering
\resizebox{1.0\textwidth}{!}{
    \begin{tikzpicture}
    \begin{axis}[
    grid=both,
    ymin=55, ymax=90,
    xmin=220, xmax=2200,
    xtick={220, 440, 880, 1320, 1760, 2200},
    extra x ticks ={1320},
    extra x tick labels={\begin{tabular}[c]{@{}c@{}} \\ (2.0\textpertenthousand{})\end{tabular} },
    xlabel=Number of sub-clouds,
    ylabel=Avg. F1 (\%),
    legend style={at={(0.7,0.3)},anchor=north}
    ]
    \addplot[mark=square, mark size=3pt, Green] plot coordinates { 
     (220,59.28)
     (440,66.17)
     (660,67.53)
     (880,71.30)
     (1100,74.09)
     (1320,73.62)
     (1540,73.84)
     (1760,76.27)
     (1980,75.97)
     (2200,77.54)
    };
    \addlegendentry{Baseline}
 
    \addplot[mark=triangle, mark size=3pt, NavyBlue] plot coordinates {
     (220,64.11)
     (440,81.90)
     (660,81.16)
     (880,83.66)
     (1100,84.43)
     (1320,85.20)
     (1540,85.01)
     (1760,84.55)
     (1980,85.48)
     (2200,86.89)
    };
    \addlegendentry{Ours}

    \addplot[dashed, YellowOrange,line width=2pt, domain=220:2200] {86.26};
    \addlegendentry{Full supervision}
    \end{axis}
    \end{tikzpicture}
    }
\end{subfigure}

\caption{The impact of the increased number of sub-cloud labels on OA and Avg. F1 for Semantic3D dataset.}
\label{fig:s3d_line}
\end{figure*}

\subsubsection{Paris-Lille-3D dataset}
\label{sec:res_res_paris}

Following the same comparison strategy, we first present the incremental gain obtained using our method, as presented in Fig.~\ref{fig:paris_line}. Compared to the baseline, the OA and average F1 scores were considerably improved by our method. However, it can be seen that OA is fairly high when trained only with initial sub-clouds. In contrast, average F1 score shows rapid growth during the early stages. This indicates dominant categories are already well classified at the beginning, while model is under-fitted for marginal ones due to the lack of adequate annotations. A rapid rise in accuracy can be observed for results of our method, attaining 90.3\% average F1 score with 1,500 sub-clouds, which accounts for 2.1\textpertenthousand{} of total labels. We present a classification map using 1,500 sub-clouds, as shown in Fig.~\ref{fig:paris_res}. With contextual enhancement, we can see that small objects such as bollard and trash can are well classified. Obvious misclassifications are shown between building and barrier, since most of points associated with these two categories exhibit similar verticality.

Several other methods are compared, and the quantitative results are listed in Table~\ref{tab:paris}. Similarly, we first analyze weakly supervised learning methods. Using 1,500 sub-clouds, our method achieves a considerable increase in the  evaluation metrics at every category compared with the baseline. Particularly, an increase of 20.4\% is observed for average F1 score. The performance of two weak supervision counterparts is still unsatisfactory even though more sub-clouds are included, and the accuracy is far below both full supervision schemes and our method. Among selected full supervision schemes, RandLA-Net underperforms with respect to average F1 score, especially ascribing the failure to categories of trash can and pedestrian. Our method achieves better average F1 score, which is attributable to the effectiveness of contextual constraints.

Furthermore, following the default benchmark setting, we submit the results on the testing set to the official ranking page\footnote{\url{https://npm3d.fr/paris-lille-3d}} for evaluation and comparison, which is presented in Table~\ref{tab:paris_test}. Intersection over Union (IoU) is utilized for evaluation, formulated as:
\begin{equation}\label{equ:IoU}
IoU = \frac{{tp}}{{tp + fp + fn}},
\end{equation}
It can be seen that using 3,200 sub-clouds ($\sim$4\textpertenthousand{} of labels), our method achieves comparable results against RandLA-Net, bur inferior to that by KPConv. A lower Avg. IoU is obtained when applying 1,600 sub-clouds ($\sim$2\textpertenthousand{} of labels), which is deviated from the performance on the validation data. Compared to the training set, the testing set shows much higher variation on scene class composition and object distribution than previously selected validation data. It might imply a limitation of our method, revealing a relatively less generalizability than full supervision schemes.

\begin{figure*}[!t]
\centering

\begin{subfigure}[b]{1.0\textwidth}
    \centering
    \includegraphics[width=0.9\linewidth]{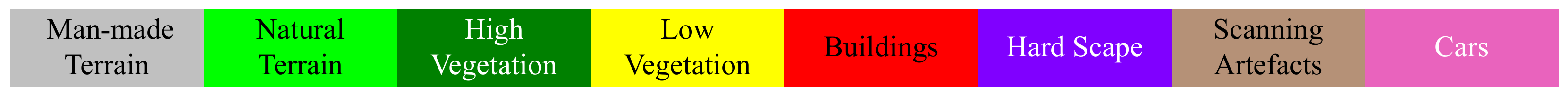}
\end{subfigure}
\hfill
\begin{subfigure}[b]{0.49\textwidth}
    \centering
    \includegraphics[width=1\linewidth]{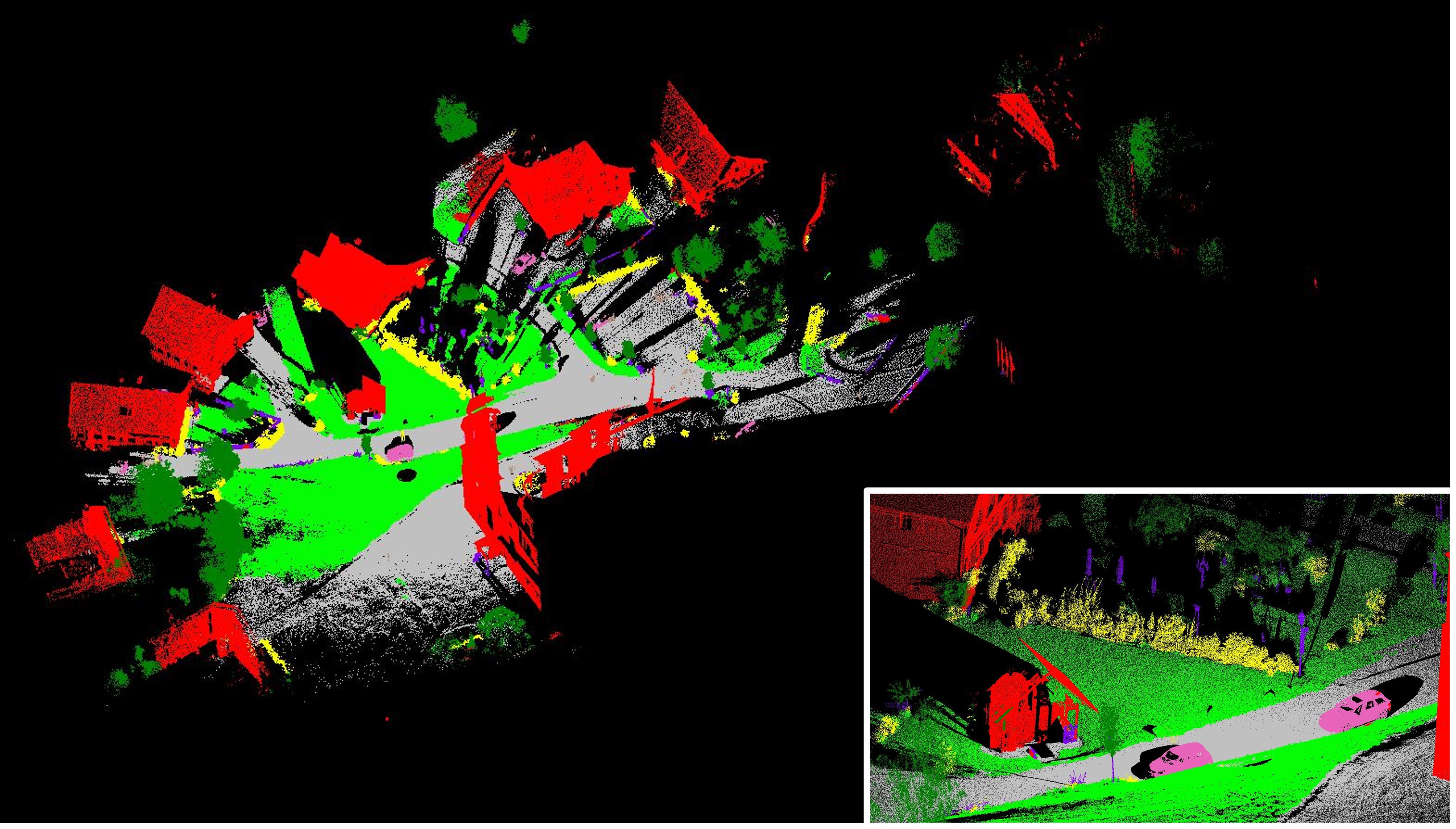}
\caption{bildstein5}
\end{subfigure}
\hfill
\begin{subfigure}[b]{0.49\textwidth}
    \centering
    \includegraphics[width=1\linewidth]{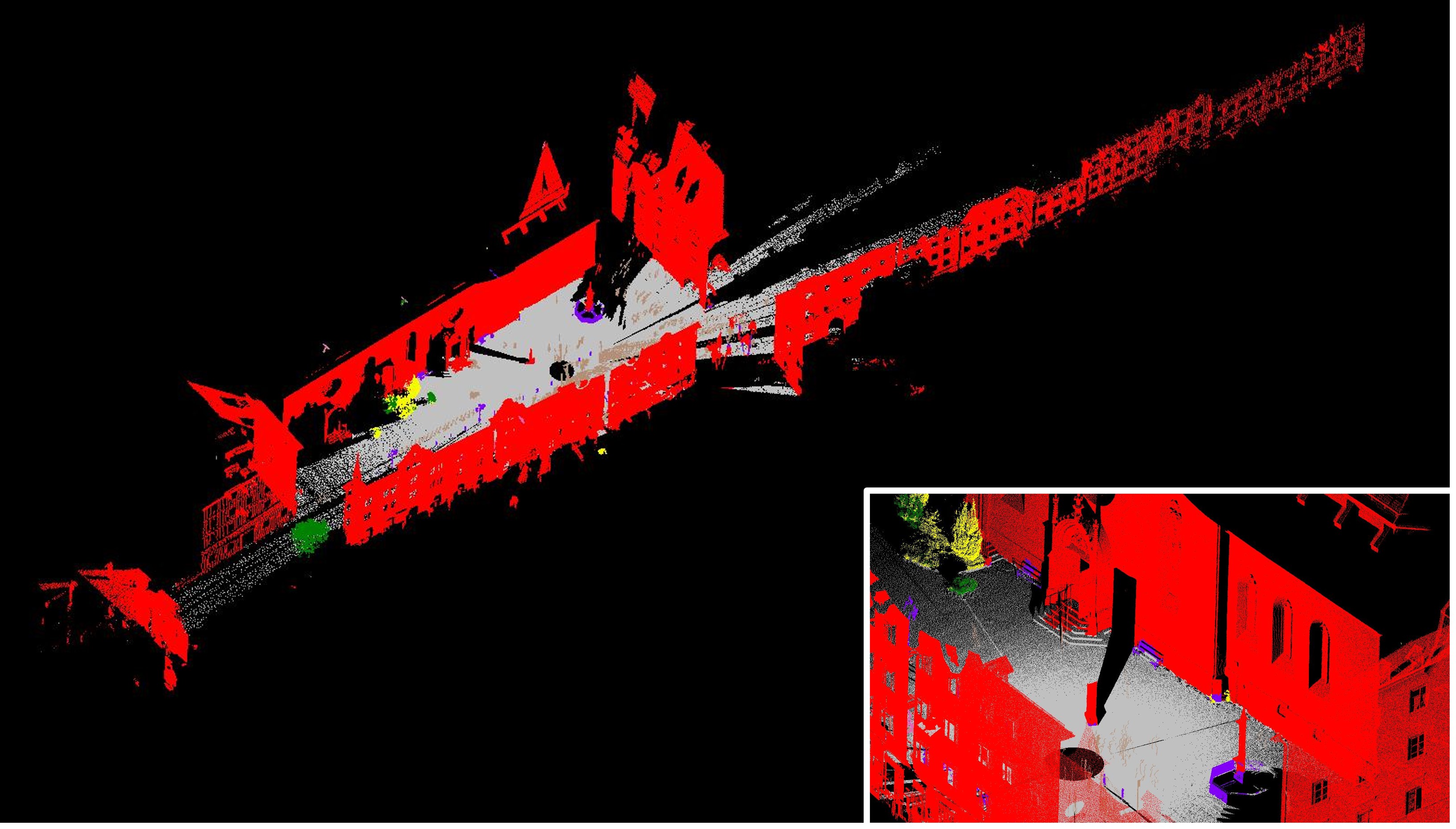}
\caption{domfountain3}
\end{subfigure}
\hfill
\begin{subfigure}[b]{0.49\textwidth}
    \centering
    \includegraphics[width=1\linewidth]{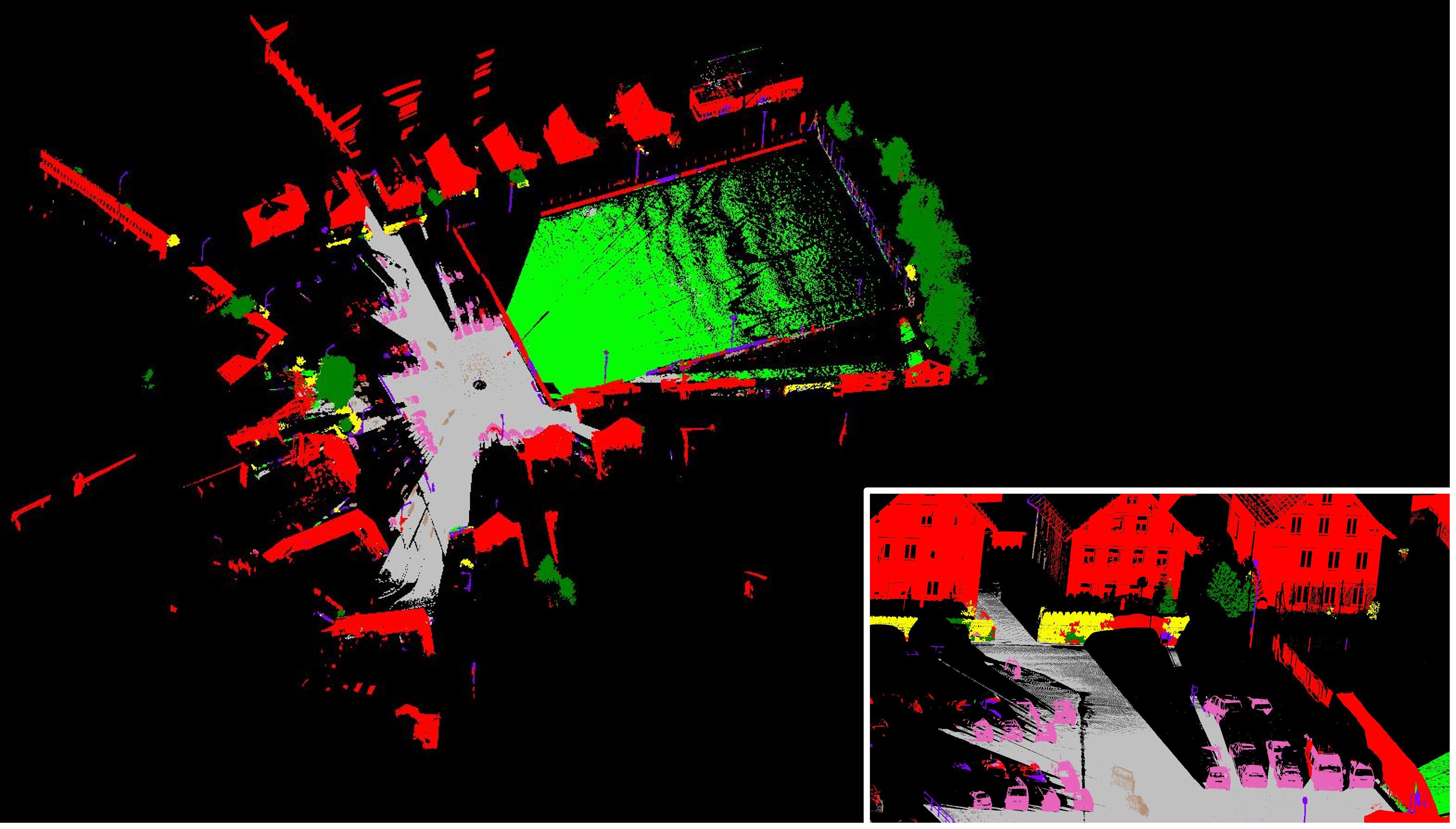}
\caption{sg27\_9}
\end{subfigure}
\hfill
\begin{subfigure}[b]{0.49\textwidth}
    \centering
    \includegraphics[width=1\linewidth]{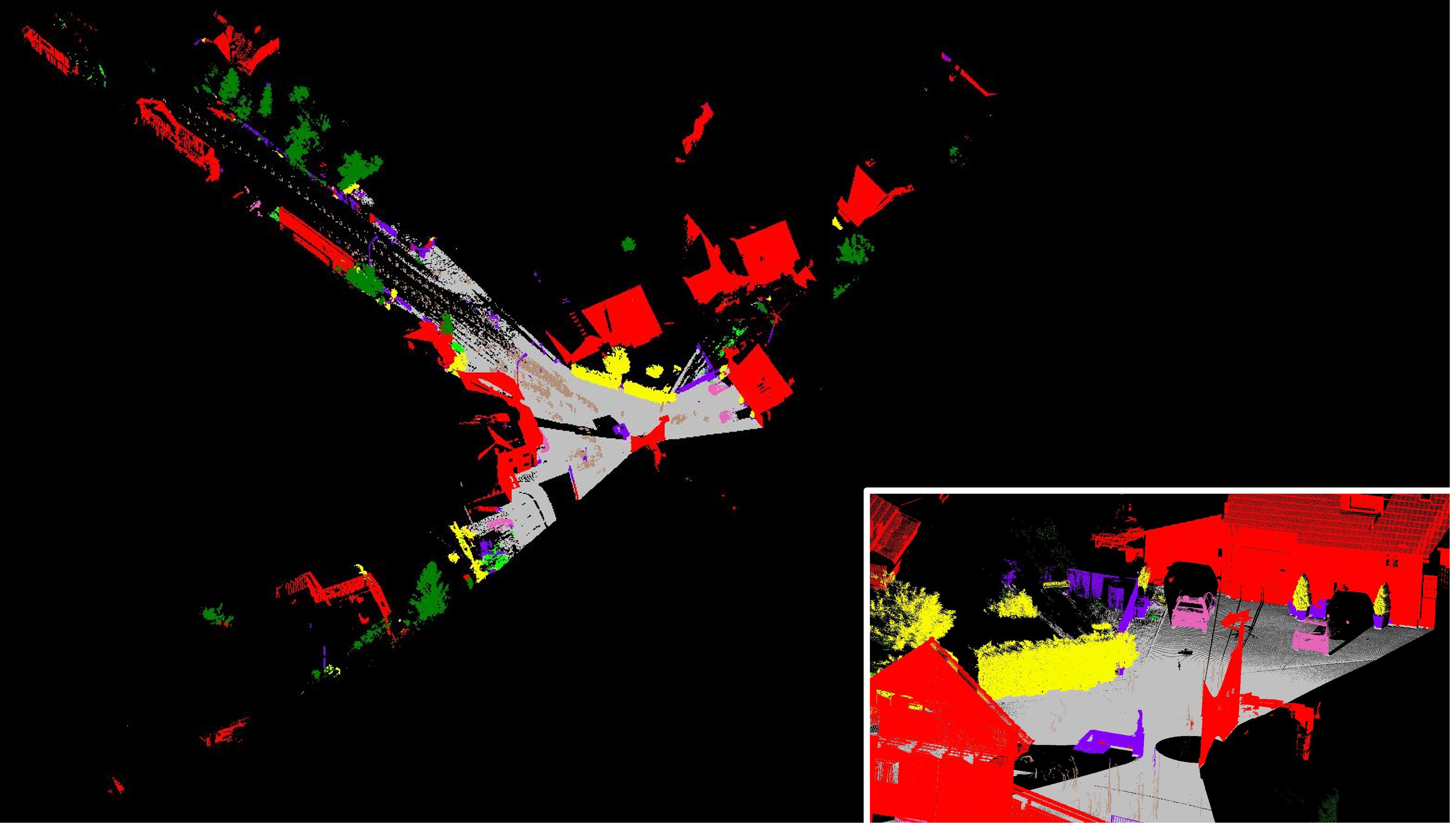}
\caption{	untermaederbrunnen3}
\end{subfigure}

\caption{Semantic3D dataset classification map using 1,320 sub-clouds.}
\label{fig:s3d_res}
\end{figure*}

\begin{table*}[t]
\caption{Comparison of full and weak supervisions for Semantic3D dataset}
\label{tab:s3d}
\centering
\small
\begin{tabularx}{\textwidth}{ccXXXXXXXXXX}
\hline
\multirow{3}{*}{Setting}  & \multirow{3}{*}{Method}  & \multicolumn{8}{c}{F1 Score}  & \multirow{3}{*}{Avg. F1}  & \multirow{3}{*}{OA} \\ 
\cline{3-10}   &  &  \begin{tabular}[c]{@{}c@{}}Man-made \\ Terrain \end{tabular}  & \begin{tabular}[c]{@{}c@{}}Natural \\ Terrain  \end{tabular}  & {\begin{tabular}[c]{@{}c@{}}High \\ Veg. \end{tabular}}  & \begin{tabular}[c]{@{}c@{}}Low \\ Veg. \end{tabular}  & Buildings  & \begin{tabular}[c]{@{}c@{}}Hard \\ Scape \end{tabular} & \begin{tabular}[c]{@{}c@{}}Scanning \\ Artefacts  \end{tabular}  & Car  &  &  \\ \hline
\multirow{2}{*}{Full Sup.}                         
& RandLA-Net  & 97.67  & 91.71  & 92.72  & 73.65  & 98.10  & 74.42  & 79.07  & 97.07  & 88.05  & 95.32  \\
& KPConv  & 98.72  & 96.83  & 95.28  & 75.60  & 97.66  & 64.91  & 67.34  & 93.78  & 86.26  & 95.46  \\ \hline
\multirow{2}{*}{\begin{tabular}[c]{@{}c@{}}Weak Sup. \\  (13,124 sc) \end{tabular}} 
& MPRM  & 92.88  & 96.34  & 86.76  &  41.77  & 95.18  & 27.42   & 18.34  & 80.84  & 67.44  & 89.68  \\
& Weak-ALS  & 91.40  & 74.83  & 86.24  & 51.81  & 94.04  & 35.23  & 39.78  & 56.91  & 66.28  & 86.36  \\ \hline
\multirow{2}{*}{\begin{tabular}[c]{@{}c@{}}Weak Sup. \\  (1,320 sc) \end{tabular}}  
& Baseline  & 95.92  & 91.50  & 88.79  & 55.30  & 94.72  & 39.57  & 46.85  & 76.32  & 73.62  & 90.56  \\
& Ours  & 98.37  & 96.27  & 94.31  & 76.65  & 96.56  & 49.87  & 78.86  & 92.69  & 85.20  & 94.61  \\ \hline
\end{tabularx}
\end{table*}

\subsubsection{Semantic3D dataset}
\label{sec:res_res_sem3d}

The performance improvement achieved by the proposed method is shown in Fig.~\ref{fig:s3d_line}, which reveals that there is an increase in the OA and average F1 score compared with baseline. Our method attains a marginally lower OA and average F1 score using 1,320 sub-clouds, which contains 2.0\textpertenthousand{} of total labels. In Fig.~\ref{fig:s3d_res}, we provide the classification results based on 1,320 sub-clouds. Owing to the high point density and distinct geometric shape of the objects, the classification results are also able to delineate the targets in the test scene with reliable boundary information. By contrast, points belong to hard scape and scanning artefacts are more likely to be wrongly classified.

The quantitative comparison results are listed in Table~\ref{tab:s3d}. Under weak label settings, our method achieves a considerable increase in OA and F1 scores for all categories compared with the baseline. Due to distinct data characteristics of terrestrial laser scanning, much more sub-clouds are queried by other two weakly supervised methods. Nevertheless, low accuracy is still present on evaluation metrics. Based on results using three datsets, we argue that current methods which purely rely on scene-level labels fail to achieve acceptable results. For marginal categories, these methods are completely useless. Moreover, using much less labels, even our baseline outperforms these two methods, showing the superiority of OCOC annotation.

\subsection{Ablation study}
\label{sec:res_ablation}

We evaluate the effectiveness of both weak supervision and active learning mechanisms proposed in our framework, respectively. Note that, when one module is under analysis, the other one remains activated. 

\begin{figure*}[!t]
\centering
    \begin{subfigure}[b]{0.32\textwidth}
    \centering
    
    \resizebox{1.0\textwidth}{!}{
    \begin{tikzpicture}
    \begin{axis}[
    grid=both,
    ymin=55, ymax=80,
    xmin=150, xmax=1500,
    xtick={150, 300, 600, 900, 1200, 1500},
    xlabel=Number of sub-clouds, 
    ylabel=Avg. F1 (\%), 
    legend style={at={(0.7,0.3)},anchor=north}
    ]
    \addplot[mark=square, mark size=3pt, OliveGreen] plot coordinates 
    { 
     (150,57.63)
     (300,60.94)
     (450,62.67)
     (600,64.76)
     (750,66.16)
     (900,69.261)
     (1050,69.83)
     (1200,69.66)
     (1350,71.46)
     (1500,71.46)
    };
    \addlegendentry{wo CC\&PL}
    
    \addplot[mark=o, mark size=3pt, red] plot coordinates 
    { 
     (150,62.50)
     (300,64.40)
     (450,67.58)
     (600,70.56)
     (750,72.65)
     (900,72.05)
     (1050,71.80)
     (1200,74.35)
     (1350,75.29)
     (1500,75.97)
    };
    \addlegendentry{w CC}
 
    \addplot[mark=triangle, mark size=3pt, NavyBlue] plot coordinates {
     (150,62.50)
     (300,66.46)
     (450,70.16)
     (600,71.40)
     (750,74.19)
     (900,74.65)
     (1050,75.40)
     (1200,75.00)
     (1350,75.56)
     (1500,75.72)
    };
    \addlegendentry{w CC\&PL (ours)}

    \end{axis}
    \end{tikzpicture}
    }
    \caption{H3D dataset}
    \end{subfigure}
    \hfill
    \begin{subfigure}[b]{0.32\textwidth}
    \centering
    
    \resizebox{1.0\textwidth}{!}{
    \begin{tikzpicture}
    \begin{axis}[
    grid=both,
    ymin=60, ymax=95,
    xmin=300, xmax=3000,
    xtick={300, 600, 1200, 1800, 2400, 3000},
    xlabel=Number of sub-clouds, 
    ylabel=Avg. F1 (\%), 
    legend style={at={(0.7,0.3)},anchor=north}
    ]
    \addplot[mark=square, mark size=3pt, OliveGreen] plot coordinates { 
     (300,61.52)
     (600,66.78)
     (800,68.79)
     (1200,70.55)
     (1500,74.34)
     (1800,79.40)
     (2100,76.36)
     (2400,78.36)
     (2700,81.01)
     (3000,81.48)
    };
    \addlegendentry{wo CC\&PL}
    
    \addplot[mark=o, mark size=3pt, red] plot coordinates { 
     (300,70.50)
     (600,74.24)
     (800,85.49)
     (1200,87.66)
     (1500,84.95)
     (1800,86.15)
     (2100,87.53)
     (2400,86.88)
     (2700,88.83)
     (3000,88.11)
    };
    \addlegendentry{w CC}
 
    \addplot[mark=triangle, mark size=3pt, NavyBlue] plot coordinates {
     (300,70.50)
     (600,80.54)
     (800,86.07)
     (1200,87.26)
     (1500,90.30)
     (1800,89.16)
     (2100,89.91)
     (2400,90.11)
     (2700,89.12)
     (3000,89.85)
    };
    \addlegendentry{w CC\&PL (ours)}
    
    \end{axis}

    \end{tikzpicture}
    }
    \caption{Paris3D dataset}
    \end{subfigure}
    \hfill
    \begin{subfigure}[b]{0.32\textwidth}
    \centering
    
    \resizebox{1.0\textwidth}{!}{
    \begin{tikzpicture}
    \begin{axis}[
    grid=both,
    ymin=60, ymax=90,
    xmin=220, xmax=2200,
    xtick={220, 440, 880, 1320, 1760, 2200},
    xlabel=Number of sub-clouds, 
    ylabel=Avg. F1 (\%), 
    legend style={at={(0.7,0.3)},anchor=north}
    ]
    \addplot[mark=square, mark size=3pt, OliveGreen] plot coordinates { 
     (220,59.28)
     (440,64.06)
     (660,68.33)
     (880,69.36)
     (1100,73.93)
     (1320,75.24)
     (1540,76.84)
     (1760,76.13)
     (1980,79.72)
     (2200,79.21)
    };
    \addlegendentry{wo CC\&PL}
    
    \addplot[mark=o, mark size=3pt, red] plot coordinates { 
     (220,64.11)
     (440,77.15)
     (660,82.59)
     (880,83.04)
     (1100,81.97)
     (1320,83.14)
     (1540,82.98)
     (1760,85.02)
     (1980,85.57)
     (2200,86.09)
    };
    \addlegendentry{w CC}
 
    \addplot[mark=triangle, mark size=3pt, NavyBlue] plot coordinates {
     (220,64.11)
     (440,81.90)
     (660,81.16)
     (880,83.66)
     (1100,84.43)
     (1320,85.20)
     (1540,85.01)
     (1760,84.55)
     (1980,85.48)
     (2200,86.89)
    };
    \addlegendentry{w CC\&PL (ours)}
    
    \end{axis}
    \end{tikzpicture}
    }
    \caption{Semantic3D dataset}
    \end{subfigure}

\caption{Effectiveness of weakly supervised method on three datasets. We refer to CC and PL as contextual constraint and pseudo labeling, respectively.}
\label{fig:wsl_res}

\end{figure*}
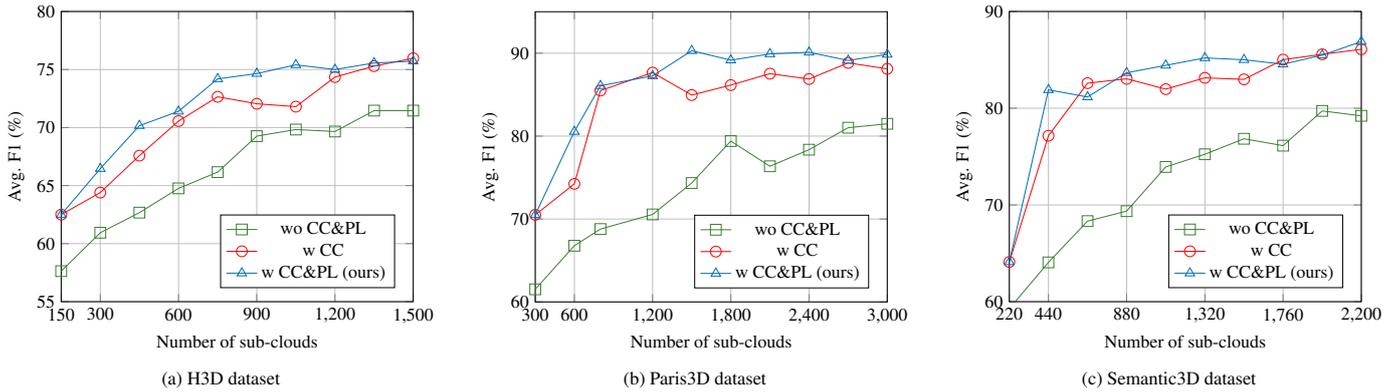

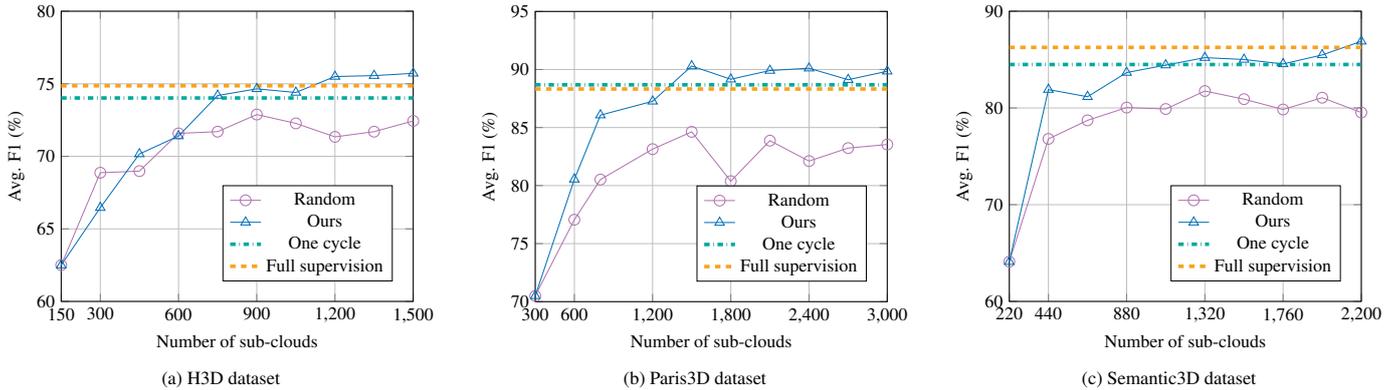
\begin{figure*}[!t]
\centering
    \begin{subfigure}[b]{0.32\textwidth}
    \centering
    
    \resizebox{1.0\textwidth}{!}{
    \begin{tikzpicture}
    \begin{axis}[
    grid=both,
    ymin=60, ymax=80,
    xmin=150, xmax=1500,
    xtick={150, 300, 600, 900, 1200, 1500},
    xlabel=Number of sub-clouds, 
    ylabel=Avg. F1 (\%), 
    legend style={at={(0.7,0.4)},anchor=north}
    ]
    
    \addplot[mark=o, mark size=3pt, Orchid] plot coordinates { 
     (150,62.50)
     (300,68.87)
     (450,68.98)
     (600,71.58)
     (750,71.70)
     (900,72.88)
     (1050,72.27)
     (1200,71.34)
     (1350,71.70)
     (1500,72.44)
    };
    \addlegendentry{Random}
 
    \addplot[mark=triangle, mark size=3pt, NavyBlue] plot coordinates {
     (150,62.50)
     (300,66.46)
     (450,70.16)
     (600,71.40)
     (750,74.19)
     (900,74.65)
     (1050,74.40)
     (1200,75.50)
     (1350,75.56)
     (1500,75.72)
    };
    \addlegendentry{Ours}

    \addplot[dash dot, Emerald, line width=2pt, domain=150:1500] {74.02};
    \addlegendentry{One cycle}
    
    \addplot[dashed, YellowOrange, line width=2pt, domain=150:1500] {74.86};
    \addlegendentry{Full supervision}

    \end{axis}
    \end{tikzpicture}
    }
    \caption{H3D dataset}
    \end{subfigure}
    \hfill
    \begin{subfigure}[b]{0.32\textwidth}
    \centering
    
    \resizebox{1.0\textwidth}{!}{
    \begin{tikzpicture}
    \begin{axis}[
    grid=both,
    ymin=70, ymax=95,
    xmin=300, xmax=3000,
    xtick={300, 600, 1200, 1800, 2400, 3000},
    xlabel=Number of sub-clouds, 
    ylabel=Avg. F1 (\%), 
    legend style={at={(0.7,0.4)},anchor=north}
    ]
    \addplot[mark=o, mark size=3pt, Orchid] plot coordinates { 
     (300,70.50)
     (600,77.06)
     (800,80.52)
     (1200,83.14)
     (1500,84.63)
     (1800,80.38)
     (2100,83.87)
     (2400,82.11)
     (2700,83.23)
     (3000,83.54)
    };
    \addlegendentry{Random}
 
    \addplot[mark=triangle, mark size=3pt, NavyBlue] plot coordinates {
     (300,70.50)
     (600,80.54)
     (800,86.07)
     (1200,87.26)
     (1500,90.30)
     (1800,89.16)
     (2100,89.91)
     (2400,90.11)
     (2700,89.12)
     (3000,89.85)
    };
    \addlegendentry{Ours}

    \addplot[dash dot, Emerald, line width=2pt, domain=300:3000] {88.68};
    \addlegendentry{One cycle}
    
    \addplot[dashed, YellowOrange,line width=2pt, domain=300:3000] {88.32};
    \addlegendentry{Full supervision}
    
    \end{axis}

    \end{tikzpicture}
    }
    \caption{Paris3D dataset}
    \end{subfigure}
    \hfill
    \begin{subfigure}[b]{0.32\textwidth}
    \centering
    
    \resizebox{1.0\textwidth}{!}{
    \begin{tikzpicture}
    \begin{axis}[
    grid=both,
    ymin=60, ymax=90,
    xmin=220, xmax=2200,
    xtick={220, 440, 880, 1320, 1760, 2200},
    xlabel=Number of sub-clouds, 
    ylabel=Avg. F1 (\%), 
    legend style={at={(0.7,0.4)},anchor=north}
    ]
    \addplot[mark=o, mark size=3pt, Orchid] plot coordinates { 
     (220,64.11)
     (440,76.80)
     (660,78.72)
     (880,80.04)
     (1100,79.89)
     (1320,81.75)
     (1540,80.91)
     (1760,79.84)
     (1980,81.06)
     (2200,79.53)
    };
    \addlegendentry{Random}
 
    \addplot[mark=triangle, mark size=3pt, NavyBlue] plot coordinates {
     (220,64.11)
     (440,81.90)
     (660,81.16)
     (880,83.66)
     (1100,84.43)
     (1320,85.20)
     (1540,85.01)
     (1760,84.55)
     (1980,85.48)
     (2200,86.89)
    };
    \addlegendentry{Ours}

    \addplot[dash dot, Emerald, line width=2pt, domain=220:2200] {84.49};
    \addlegendentry{One cycle}

    \addplot[dashed, YellowOrange,line width=2pt, domain=220:2200] {86.26};
    \addlegendentry{Full supervision}
    
    \end{axis}
    \end{tikzpicture}
    }
    \caption{Semantic3D dataset}
    \end{subfigure}

\caption{Effectiveness analysis of active learning. We refer to one cycle as training the model only once with maximum number of sub-clouds. Random mode stands for randomly selecting sub-clouds for implementing OCOC.}
\label{fig:al_res}
\end{figure*}

\subsubsection{Effectiveness of weakly supervised strategy}
\label{sec:res_ablation_ws}

We analyze the effectiveness of pseudo labels and contextual constraints which build the key components within our weakly supervised strategy, which is presented in Fig.~\ref{fig:wsl_res}. Without both two modules, poor results are produced on all three datasets. With contextual constraint incorporated, we can see that the average F1 score has considerably increased for all datasets during every training cycle. For Paris3D and Semantic3D datasets, it largely enhances performance already starting from early training stages, which is increased by approximately 15\%. There is also a noticeable rise of 5\% for H3D datset. Moreover, even better results are achieved by integrating context-aware pseudo labels. Although satisfactory results can be obtained at the end of the iterative training process without pseudo labeling, reliable supplementary supervisory signals from pseudo labeling expedite the model performance boost, which enables a more labeling and training efficient mode. It shows the method combining two modules achieves the best result for all datasets, demonstrating the effectiveness of our method.

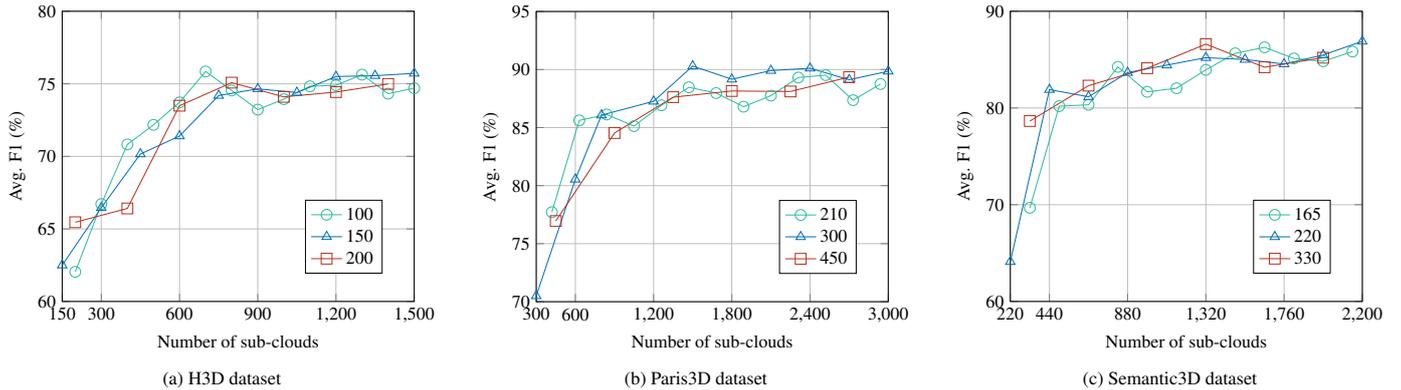
\begin{figure*}[t]
\centering
    \begin{subfigure}[b]{0.32\textwidth}
    \centering
    
    \resizebox{1.0\textwidth}{!}{
    \begin{tikzpicture}
    \begin{axis}[
    grid=both,
    ymin=60, ymax=80,
    xmin=150, xmax=1500,
    xtick={150, 300, 600, 900, 1200, 1500},
    xlabel=Number of sub-clouds, 
    ylabel=Avg. F1 (\%), 
    legend style={at={(0.8,0.35)},anchor=north}
    ]
    \addplot[mark=o, mark size=3pt, SeaGreen] plot coordinates { 
     (200,62.04)
     (300,66.71)
     (400,70.82)
     (500,72.17)
     (600,73.72)
     (700,75.85)
     (800,74.54)
     (900,73.22)
     (1000,73.94)
     (1100,74.83)
     (1200,74.90)
     (1300,75.63)
     (1400,74.32)
     (1500,74.70)
    };
    \addlegendentry{100}
 
    \addplot[mark=triangle, mark size=3pt, NavyBlue] plot coordinates {
     (150,62.50)
     (300,66.46)
     (450,70.16)
     (600,71.40)
     (750,74.19)
     (900,74.65)
     (1050,74.40)
     (1200,75.50)
     (1350,75.56)
     (1500,75.72)
    };
    \addlegendentry{150}
    
    \addplot[mark=square, mark size=3pt, BrickRed] plot coordinates {
     (200,65.46)
     (400,66.41)
     (600,73.49)
     (800,75.08)
     (1000,74.10)
     (1200,74.43)
     (1400,74.99)
    };
    \addlegendentry{200}

    \end{axis}
    \end{tikzpicture}
    }
    \caption{H3D dataset}
    \end{subfigure}
    \hfill
    \begin{subfigure}[b]{0.32\textwidth}
    \centering
    
    \resizebox{1.0\textwidth}{!}{
    \begin{tikzpicture}
    \begin{axis}[
    grid=both,
    ymin=70, ymax=95,
    xmin=300, xmax=3000,
    xtick={300, 600, 1200, 1800, 2400, 3000},
    xlabel=Number of sub-clouds, 
    ylabel=Avg. F1 (\%), 
    legend style={at={(0.8,0.35)},anchor=north}
    ]
    
    \addplot[mark=o, mark size=3pt, SeaGreen] plot coordinates { 
     (420,77.71)
     (630,85.62)
     (840,86.13)
     (1050,85.13)
     (1260,86.92)
     (1470,88.47)
     (1680,87.97)
     (1890,86.80)
     (2100,87.75)
     (2310,89.31)
     (2520,89.52)
     (2730,87.36)
     (2940,88.75)
    };
    \addlegendentry{210}
 
    \addplot[mark=triangle, mark size=3pt, NavyBlue] plot coordinates {
     (300,70.50)
     (600,80.54)
     (800,86.07)
     (1200,87.26)
     (1500,90.30)
     (1800,89.16)
     (2100,89.91)
     (2400,90.11)
     (2700,89.12)
     (3000,89.85)
    };
    \addlegendentry{300}
    
    \addplot[mark=square, mark size=3pt, BrickRed] plot coordinates {
     (450,76.95)
     (900,84.53)
     (1350,87.62)
     (1800,88.15)
     (2250,88.11)
     (2700,89.35)
    };
    \addlegendentry{450}

    \end{axis}

    \end{tikzpicture}
    }
    \caption{Paris3D dataset}
    \end{subfigure}
    \hfill
    \begin{subfigure}[b]{0.32\textwidth}
    \centering
    
    \resizebox{1.0\textwidth}{!}{
    \begin{tikzpicture}
    \begin{axis}[
    grid=both,
    ymin=60, ymax=90,
    xmin=220, xmax=2200,
    xtick={220, 440, 880, 1320, 1760, 2200},
    xlabel=Number of sub-clouds, 
    ylabel=Avg. F1 (\%), 
    legend style={at={(0.8,0.35)},anchor=north}
    ]
    \addplot[mark=o, mark size=3pt, SeaGreen] plot coordinates { 
     (330,69.68)
     (495,80.19)
     (660,80.33)
     (825,84.25)
     (990,81.67)
     (1155,82.04)
     (1320,83.94)
     (1485,85.65)
     (1650,86.27)
     (1815,85.12)
     (1980,84.83)
     (2145,85.83)
    };
    \addlegendentry{165}
 
    \addplot[mark=triangle, mark size=3pt, NavyBlue] plot coordinates {
     (220,64.11)
     (440,81.90)
     (660,81.16)
     (880,83.66)
     (1100,84.43)
     (1320,85.20)
     (1540,85.01)
     (1760,84.55)
     (1980,85.48)
     (2200,86.89)
    };
    \addlegendentry{220}
    
    \addplot[mark=square, mark size=3pt, BrickRed] plot coordinates {
     (330,78.65)
     (660,82.31)
     (990,84.11)
     (1320,86.60)
     (1650,84.21)
     (1980,85.20)
    };
    \addlegendentry{330}

    \end{axis}

    \end{tikzpicture}
    }
    \caption{Semantic3D dataset}
    \end{subfigure}
\caption{Influence of sample size in each training cycle}
\label{fig:sample_size}
\end{figure*}

\subsubsection{Effectiveness of active learning strategy}
\label{sec:res_ablation_al}

Regarding the active learning strategy, we compare our proposed method with other two schemes. Random scheme is random extraction of sub-clouds and OCOC during each training cycle, and one cycle scheme is to train the model only once using the same count of the sub-clouds as that from the final cycle. Please note that the proposed weakly supervised module was still applied in this ablation study. From the result in Fig.~\ref{fig:al_res}, we observe that active learning method outperforms the random selection throughout the entire training process. Obvious gaps can be observed during the late training stages. As the training proceeds, active learning seeks to identify and explore most informative samples to promote model performance. By contrast, the redundancy in randomly selected sub-clouds hinders the model to achieve more accurate results. This implies an advantage of our active learning method, which enables to mitigate the imbalanced sample issue. Since marginal categories with a small sample size are often misclassified under weak supervision, our active learning scheme help identify those points associated with most uncertain categories as informative samples, leading to annotating those points with higher probability. Thus, it could produce a more balanced weak label set, which contributes to boosting the average F1 score. An notable finding is that acceptable performance can be obtained from one cycle mode. We attribute this to the effectiveness of our weakly supervised method, as it seems capable to capture and learn effective semantic information using reasonable number of training samples. Compared to one-cycle mode, our proposed method achieves better results by means of active learning when only using half number of sub-clouds.

\subsection{Influence of sample size in each cycle}
Sample size for each training cycle is an essential hyperparameter for active learning approaches, which determines the number of human-computer interactions given a fixed annotation budget. We analyze the effectiveness of our method with three settings for the sample size, and the results are presented in Fig.~\ref{fig:sample_size}. From the line charts, we can observe that our method behaves with similar performance under all settings for the iterative training. By comparing results across different settings, a further finding is that results with less sample size for each cycle shows slightly less stability. It might be due to the reason that the sample size is too small to represent desirable data regions to cover complete label distribution. Despite this, robust performance and results can be further obtained once the size/number of sub-clouds starts to increase.

\subsection{Robustness analysis}
\label{sec:res_robust}

 The robustness of algorithms has significant influence on the reproducibility of generated results. Due to the lack of labels under weak supervision, the issue of robustness is particularly worthy studying. Since our method applies random label initialization for the first training cycle, we record average performance and corresponding standard deviation by training the model for five times. From the Table~\ref{tab:robust}, we can see that it maintains satisfactory performance within all iterations. In contrast, there is slightly higher standard deviation shown on H3D dataset. Even though the training process starts with different weak labels, our method can still leverage limited information and incrementally improve the model performance, achieving satisfactory results with a reasonable number of training cycles. It demonstrates that our active learning strategy can explore most informative samples adaptively based on currently trained model.

\begin{table}[ht]
\caption{Robustness analysis of the proposed method }
\label{tab:robust}
\centering
\small
\setlength\tabcolsep{3pt}%
\begin{tabularx}{\linewidth}{cXXXXXX}
\hline
\multirow{2}{*}{} & \multicolumn{2}{c}{H3D} & \multicolumn{2}{c}{Paris3D} & \multicolumn{2}{c}{Semantic3D} \\ 
\cline{2-7} & Avg. F1 & OA  & Avg. F1 & OA & Avg. F1 & OA  \\ \hline
Iter1  & 75.72  & 87.74  & 89.85  & 98.59  & 86.89  & 94.96  \\
Iter2  & 75.80  & 87.85  & 91.21  & 98.46  & 85.18  & 94.46  \\
Iter3  & 76.20  & 87.79  & 89.09  & 98.43  & 84.72  & 94.51  \\
Iter4  & 74.54  & 87.25  & 89.97  & 98.58  & 84.99  & 94.72  \\
Iter5  & 74.17  & 86.66  & 89.85  & 98.37  & 85.10  & 94.45  \\
\hline
\textbf{Mean}  & 75.29  & 87.46  & 89.99  & 98.49  & 85.38  & 94.62  \\
\textbf{STD}  & 0.79  & 0.45  & 0.68  & 0.01  & 0.77  & 0.20  \\  \hline
\end{tabularx}
\end{table}

\section{Conclusion}
\label{sec:con}

In this study, we investigated point cloud semantic segmentation with limited annotations and proposed an active weakly supervised framework leveraging quasi scene-level weak labels. One Class One Click (OCOC), allocating one point-level label to each of included categories in a sub-cloud, was first introduced as a new weak label format, which encompasses both scene-level and point-level semantic information. Based on it, we proposed contextual constraints and context-aware pseudo labels to enhance global feature embedding and point-wise predictions. Moreover, we incorporated active learning strategy for identifying and exploring most informative sub-clouds for corresponding OCOC annotations, which enabled a time-efficient training mode with very low labor costs. Comprehensive experiments were performed to evaluate the proposed method using three LiDAR benchmarks of different modality. Our method significantly improved OA and average F1 score compared with the baseline method. With extremely low labeling costs, competitive results are also achieved on par with other fully supervised approaches using a completely annotated training dataset. Moreover, our proposed method significantly outperforms the existing weakly supervised counterparts using scene-level labels in terms of model effectiveness and labeling efficiency. Evaluated on the H3D dataset using approximately 2.3\textpertenthousand{} of labels, our method achieved an overall accuracy of 86.46\% and an average F1 score of 74.19\%, which increased by approximately 7.52\% and 8.39\%, respectively, compared to the baseline. Even bigger performance boost can be observed for Paris3D and Semantic3D datsaets in terms of average F1 score. 

In future study, we will explore and enhance the transferability of the developed deep models. With domain adaptation between different geographical scenes with overlapped classes, present semantic knowledge could be exploited and transferred into unseen data, which further secures cost-effective labeling workload.

\section*{Acknowledgements}

This work was supported by National Natural Science Foundation of China (Project No.42171361) and the Research Grants Council of the Hong Kong Special Administrative Region, China, under Project PolyU 25211819. This work was also funded by the research project (Project Number: 2021.A6.184.21D) of the Public Policy Research Funding Scheme of The Government of the Hong Kong Special Administrative Region. This work was partially supported by The Hong Kong Polytechnic University under Projects 1-ZVN6, 1-YXAQ and Q-CDAU. The Hessigheim 3D dataset was provided by Institute for Photogrammetry, University of Stuttgart. The Paris-Lille-3D dataset was provided by Mines ParisTech, PSL Research University. The Semantic3D dataset was provided by IGP and CVG, ETH Zurich.




\bibliographystyle{elsarticle-harv} 
\bibliography{ref}





\end{document}